\documentclass[letterpaper, 10 pt, conference]{ieeeconf}  % Comment this line out if you need a4paper

\IEEEoverridecommandlockouts                              % This command is only needed if 
\usepackage{graphics} % for pdf, bitmapped graphics files
\usepackage{epsfig} % for postscript graphics files
\usepackage{mathptmx} % assumes new font selection scheme installed
\usepackage{times} % assumes new font selection scheme installed
\usepackage{amsmath} % assumes amsmath package installed
\usepackage{amssymb}  % assumes amsmath package installed
\usepackage{float}
\usepackage{hyperref}
\usepackage{subcaption}
\usepackage{graphicx}
\usepackage{multirow}
\usepackage{stfloats}
\usepackage{amsfonts}
\usepackage{siunitx}
\usepackage{algorithm}
\usepackage[noend]{algorithmic}
\usepackage{comment}
\usepackage{breqn}
\usepackage{booktabs}
\usepackage[font=footnotesize, belowskip=-10pt]{caption}

\title{\LARGE \bf
View management for lifelong visual maps
}

\author{%
Nandan Banerjee*, Ryan C. Connolly, Dimitri Lisin, Jimmy Briggs, Manjunath Narayana, and Mario E. Munich%
\thanks{*The authors are with the iRobot Corporation, Bedford, MA 01730, USA. \{nbanerjee, dlisin, jbriggs, mnarayana, mmunich\}@irobot.com, ryan.connolly@duke.edu}%
}

\begin{document}

% Dima, are you there?

\maketitle
\thispagestyle{empty}
\pagestyle{empty}

%%%%%%%%%%%%%%%%%%%%%%%%%%%%%%%%%%%%%%%%%%%%%%%%%%%%%%%%%%%%%%%%%%%%%%%%%%%%%%%%
\begin{abstract}
% introduce the concept of a landmark.
The time complexity of making observations and loop closures in a graph-based visual SLAM system is a function of the number of views stored~\cite{Eade2010,konolige2010view}. Clever algorithms, such as approximate nearest neighbor search, can make this function sub-linear. Despite this, over time the number of views can still grow to a point at which the speed and/or accuracy of the system becomes unacceptable, especially in computation- and memory-constrained SLAM systems. However, not all views are created equal. Some views are rarely observed, because they have been created in an unusual lighting condition, or from low quality images, or in a location whose appearance has changed. These views can be removed to improve the overall performance of a SLAM system. In this paper, we propose a method for pruning views in a visual SLAM system to maintain its speed and accuracy for long term use. 

\end{abstract}

% Come up with a good title for the paper

%%%%%%%%%%%%%%%%%%%%%%%%%%%%%%%%%%%%%%%%%%%%%%%%%%%%%%%%%%%%%%%%%%%%%%%%%%%%%%%%
\section{Introduction}

% need citation, cannot just say mobile robots today do this and that without proper citations (RYAN)
% Could we say something like "Although much of the technology available today is proprietary, we are unaware of commercially available robots that do not take advantage of fiducials to track their environment, or use a pre-computed static map such as the iRobot Ava"
Mobile robots capable of mapping their environment using SLAM are increasingly common today, not only in research labs and warehouses, but also in homes and businesses. Often such robots create a map once, and then use it to perform their tasks \cite{ava500}. A static map is sufficient for highly controlled environments such as a warehouse, or spartan environments such as hotel hallways. In more common household environments, dynamic conditions arise due to variations in lighting, movement of furniture, and appearance of clutter. In time, a static map no longer represents the current state of the environment, leading to degradation or failure in localization.

Ideally, a robot should detect changes in its environment and update its map accordingly. Maintaining an up-to-date representation of the space allows the robot to estimate its location more accurately and to make better navigation decisions. We refer to the maps being continuously updated by the robot as {\em lifelong visual maps}~\cite{konolige2009towards}.

\begin{figure}[h!]
    \centering
    \includegraphics[width=0.4\textwidth]{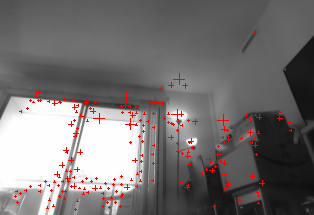}
    \caption{A sample view from an environment as seen by our robot with the red crosshairs indicating detected keypoints.}
    \label{fig:sample_view}
\end{figure}

In the SLAM literature, the term {\em view} refers to a uniquely identifiable scene in the world whose location can be estimated by the robot using its sensors. In the context of vision-based SLAM systems, the sensor is a camera, and a view is a representation of a visual scene (see Fig.~\ref{fig:sample_view}) that can be observed by the robot to estimate its location and orientation. 

The main contribution of this paper is an efficient view management algorithm for resource-constrained visual SLAM systems. The algorithm selects views that can be removed from the map without sacrificing localization performance. View removal is based on observation statistics collected over time. The algorithm has been developed for the monocular visual SLAM system by Eade at al.~\cite{Eade2010}; however, it can be easily adapted to fit any graph-based SLAM system containing a notion of views.

\begin{figure}[b]
    \centering
    \includegraphics[width=0.4\textwidth]{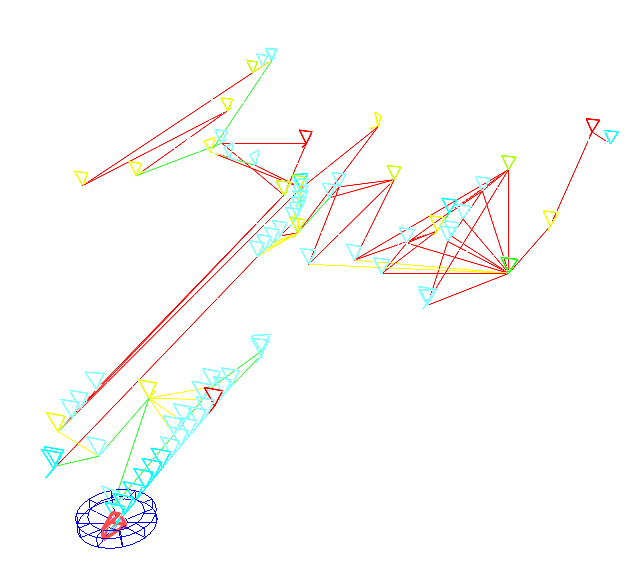}
    \caption{The graph as created by the robot over a 10-minute run, where the cyan $\nabla$ represents pose nodes, the red, yellow, and green $\nabla$ represents view nodes that are new, slightly observed, and frequently observed respectively. The yellow edges are observation edges, the green ones are combined observation edges, and the red ones are combined observation and odometry measurement edges.}
    \label{fig:graphslam_running}
\end{figure}

In our system, we represent the map of the robot's environment as a pose graph~\cite{Eade2010} where each node represents either a robot pose or a view, and each edge represents a measurement either between a pair of robot poses or between a robot pose and a view (see Fig.~\ref{fig:graphslam_running}). Pose-to-pose edges give a robot trajectory estimate that may drift over time due to measurement errors from dead reckoning sensors. Pose-to-view edges create loop closures in the graph \cite{Eade2010}. These loop closures allow the SLAM system to minimize accumulated measurement errors with a graph optimization algorithm (e. g. \cite{poseGraph, factorGraphs}) thus correcting for the drift.

Conceptually our SLAM system can be divided into the {\em back end}, responsible for constructing and optimizing the graph, and the {\em front end} responsible for creating and recognizing the views. A view is created by selecting a pair of video frames, finding point correspondences between them, triangulating the resulting pairs of points, and refining the 3D points using non-linear optimization to minimize the re-projection errors.

The back end's most expensive operation is the nonlinear graph optimization, whose complexity is a function of the number of nodes and edges in the graph. These in turn are a function of time rather than space. As the robot moves, the graph grows, and the cost of determining the robot's pose can become prohibitive. To counteract this problem, graph sparsification techniques that periodically remove nodes and edges have been developed \cite{Eade2010, Duy2018}.

In the front end, most of the computational cost comes from observing existing views. Creating a new view incurs a constant cost, because it does not use information about existing views. In contrast, view observation requires the robot to determine if an image captured by the camera matches one of the known views. In our system, we select a fixed number of likely candidates from the pool of all views using a candidate selection algorithm, and then compare these candidates to the current image. Thus, the complexity of the candidate selection method during the process of view observation is a function of the total number of views in the system.

\begin{figure}[h!]
    \centering
    \includegraphics[width=0.5\textwidth]{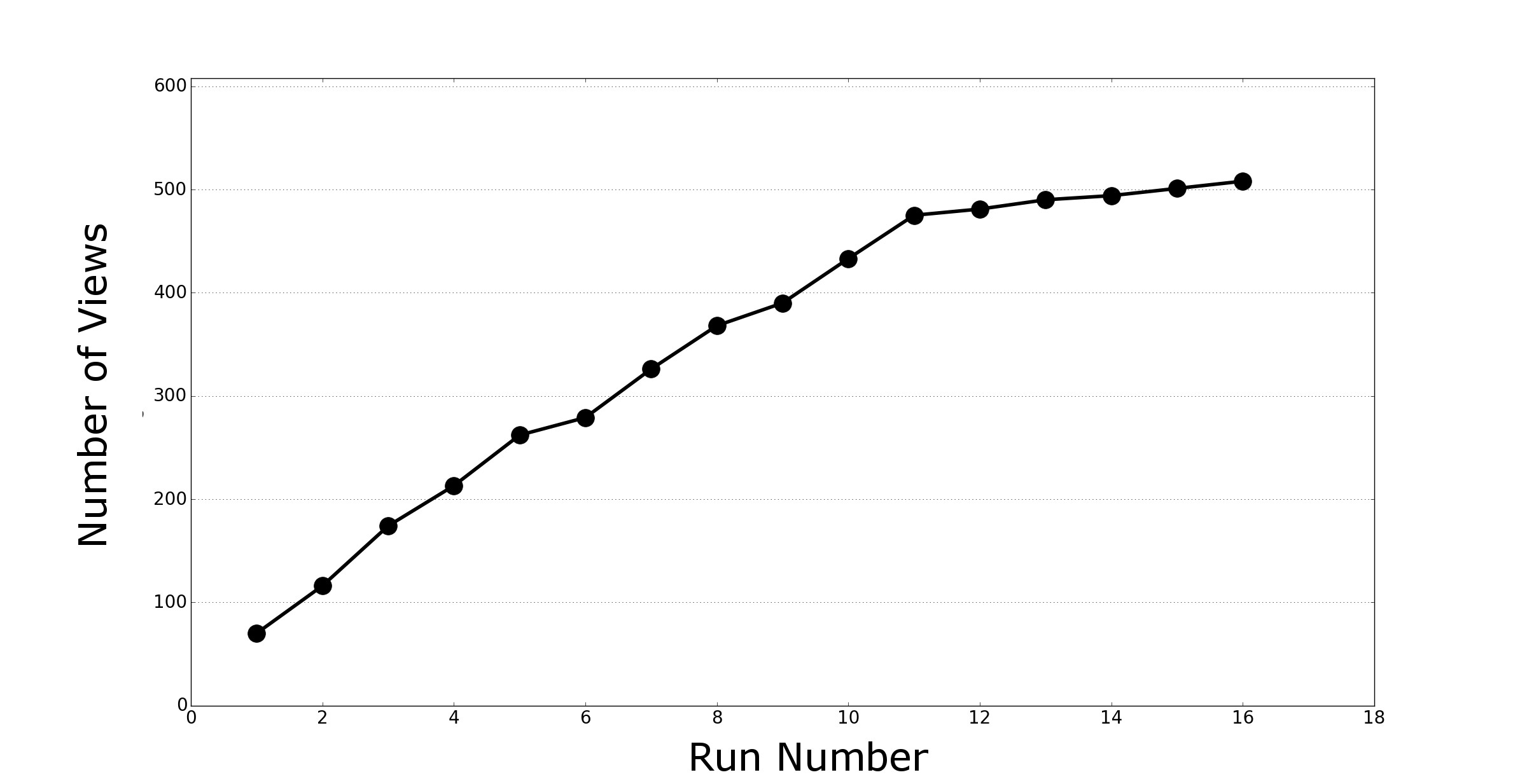}
    \caption{The number of views in a visual SLAM system grows over time. The graph shows the number of views increasing as the robot runs in the same environment several times under different lighting conditions. Eventually, too many views will degrade the speed and accuracy of the system.}
    \label{fig:landmark_growth_over_time.}
\end{figure}

\begin{figure}[h!]
    \centering
    \includegraphics[width=0.35\textwidth]{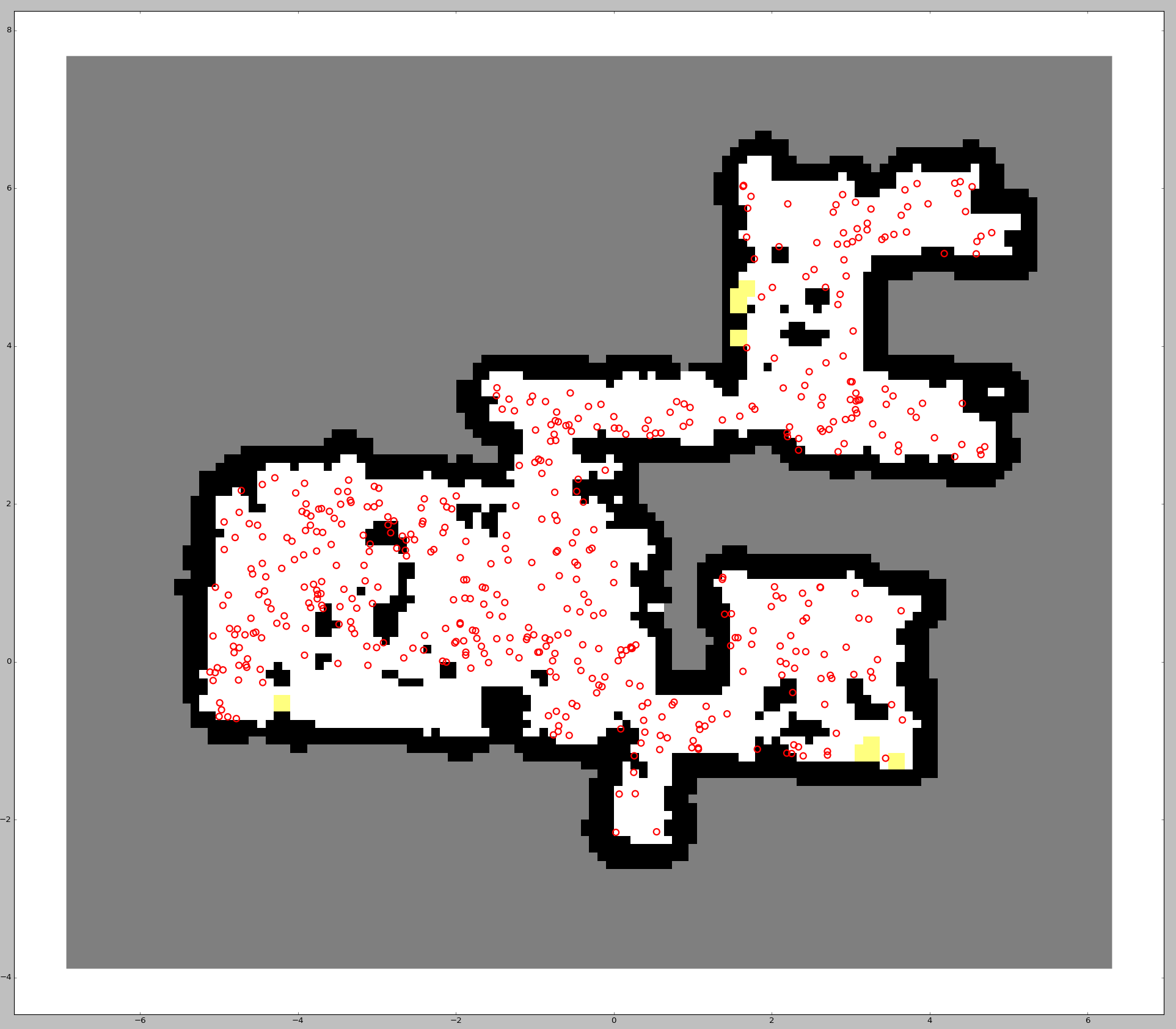}
    \caption{Map of an environment (around 500 $ft^2$) without view management after 20 runs -- the distribution of views (represented in red) in the map is very dense.}
    \label{fig:env_B}
\end{figure}

If the robot operates for long periods of time in the same space, it might continue to create new views because of changes in the appearance of its surroundings, such as different lighting conditions or moving furniture (Fig.~\ref{fig:landmark_growth_over_time.}).  Eventually, the distribution of the views can become very dense (Fig.~\ref{fig:env_B}), which can make view observation unacceptably slow, and degrade the robot's performance, rather than improve the localization accuracy. Unfortunately, graph pruning only removes pose nodes, it does not limit the growth of the number of views over time.

% should have an image of the space where these views were made, maybe overlay the views over the map

% should also probably have an image of the decrease in the total number of observations with more views in the system 

% pictures on view growth over time ... more time to make observations ... decrease in number of observations

\section{Related Work}
%nandan \& emily \& ryan

Reduction of computation and memory devoted to view recognition has been an ongoing goal of the SLAM community for some time. In this section, we explore some of the more successful approaches and compare them to our algorithm.

Eade et al. attempt to control the complexity of view recognition by using an approximate nearest-neighbor algorithm~\cite{lowe-kd-trees}. Instead of comparing the camera image to all existing views, it pools the visual feature descriptors from the views into a tree structure for approximate nearest neighbor matching, and uses a voting scheme to select a short list of candidates. This approach is logarithmic in the number of the view features, and sub-linear in the number of views. Nevertheless, if the number of views is allowed to grow unabated, eventually the cost of view recognition becomes prohibitive for a resource-constrained robot platform. It is therefore necessary for the SLAM system to keep the number of views below a maximum determined by the hardware capabilities.

View clustering techniques use various heuristics to prune views which are likely to duplicate information. Konolige and Bowman~\cite{konolige2009towards} introduced this class of methods with a ``least recently used heuristic'' to delete views. Hochdorfer et al. explored clustering algorithms to reduce the co-observation of features in multiple views ~\cite{hochdorfer}. These early works decreased the number of views stored without significantly affecting localization performance, but were sensitive to changes in lighting and movement of clutter. Our algorithm maintains performance even with these variations. Other heuristics developed include approaches which remove views unlikely to survive the ANN search \cite{Hartmann}, filter by estimates of distinctiveness, detectability, and repeatability~\cite{buoncompagni2015saliency}, or learn which features are most likely to perform well, and remove the others~\cite{erasingbadmemories}.

M\"uhlfellner et al.~\cite{summary_maps_for_lifelong_visual_localization} proposed a two-phase algorithm known as \textit{Summary Maps}. In the first phase, high quality views are selected in an offline processing step. In the second phase, the resultant subset of views (the summary map) is sent back to the navigation system and used for subsequent SLAM runs. B\"urki et al. extended this approach to match loaded summary maps to lighting conditions~\cite{BuerkiAppearance}, and Dymczyk et al. explored linear and quadratic programming approaches for the offline computation step~\cite{keepitbrief}. While these techniques do speed up computation in the vision front-end, they all require expensive offline computation. Often this computation is outsourced from an agent to a separate server. On resource-constrained, independent platforms, however, this offline computation is prohibitively expensive. Our method can manage views continuously on the agent itself during SLAM and thus can work without requiring a connection to a server.

Other work on lifelong localization and mapping addresses outdoor autonomous systems with high compute power, such as those used by self-driving cars~\cite{Churchill,Churchilltwo}. Because of this high computational capacity, such systems are not overly concerned with reducing the size of the map.

\section{Problem Statement}
A mobile robot equipped with a visual SLAM system starts building a new map at the beginning of a run, and maintains localization by observing the map's views. Up to a point, having more views improves the robot's localization accuracy, by increasing the likelihood of observation. However, the number of views cannot grow without bound. There is a critical number of views, beyond which the speed or accuracy of retrieving a matching view becomes unacceptable, or which simply exceeds the available memory capacity. This maximum number of views must be determined experimentally by testing that the system runs at an acceptable frame rate, leaves enough CPU cycles for other tasks needed by the robot, and, of course, does not crash because it ran out of memory.

In a confined space where the lighting does not change much, the SLAM system may stop creating new views before their number reaches the maximum imposed by the hardware, but in a large space where lighting conditions can vary widely, the system is likely to reach the maximum number of views eventually, especially if it saves the map at the end of a run, and re-uses it for the next run. Therefore, we need a method for pruning the views to keep their number under the limit.

\section{Method}
\label{sec:method}

In this section, we describe our proposed algorithm for identifying views that are to be pruned. First, we define the terminology used in our visual SLAM system. Then we discuss the properties of our system that we would like to preserve after the pruning of views. Finally, we explain the metrics used for determining the quality of a view and explain our proposed algorithm for managing views. We use the terms {\em management} and {\em pruning} interchangeably in this paper.

The SLAM graph can consist of multiple disjoint sub-graphs, called {\em components}, which ideally represent different environments. A map is defined as a component of the graph with all its associated views and occupancy information. Note that each component has its own coordinate system.

When the robot begins a run without any prior knowledge of its location, it creates a new component. It may also have one or more components in its SLAM graph, which were saved during the previous runs. Let us denote the newly created component as $A$. If there is a component $B$ in the graph, corresponding to the robot's current environment, then the robot is able to merge $A$ and $B$ after observing some minimum number of views from $B$. We call this event a {\em relocalization} into a map. 

There are several properties of the SLAM system that we must preserve throughout the view pruning process. First, despite removing some of the views from a component, the robot must still reliably observe enough of the remaining views to relocalize into that component. The appearance of any given environment can change drastically between runs, because of lighting changes and moved objects. These changes can make it impossible for the robot to observe some of the views during certain runs. For example, a view that was created during daylight hours might not be observable in the evening, when the electric lights are on. Other views may be observable in both lighting conditions. It is critical to keep more of the views, which were observed in multiple runs, which are more likely to allow the robot to relocalize despite the appearance changes. On the other hand, removing the views that have not been observed in a long time provides a way for the SLAM system to forget visual scenes that are no longer observable, such as when furniture has been moved.

Along with preserving the views observable across different runs, view pruning should also attempt to keep the distribution of views across a component as uniform as possible.  Otherwise, there may be areas in the environment, where a robot can travel for too long without observing a view, and thus accumulate localization error. We enforce the uniformity of the view distribution by pruning only those views which neighbor other views of similar orientation.

We now present our algorithm for selecting a set of views in a component that can be pruned without negatively affecting relocalization, or the uniform spatial distribution of views. First, we describe the parameters of the algorithm, then the score calculation, and finally the algorithm itself.

\subsection{Parameters}

The following are the parameters of our algorithm:
\begin{itemize}
    \item MIN\_VIEWS : Minimum number of views required before the pruning algorithm is run. The value of this parameter should be determined experimentally, to make sure that the SLAM system runs at the required frame rate, does not crash, and leaves enough cycles for other tasks.
    \item NN\_THRESHOLD : Nearest neighbor threshold, if the total number of views nearby is less than the threshold, the view will not be marked for deletion.
    \item NN\_VOXEL\_SIZE : 3-element vector (x, y, and $\theta$) used for constraining the search space while finding neighboring views.
    \item SCORE\_THRESHOLD : If the view score is less than this threshold, the view will be deleted.
    \item $W_{1} $ : Used to weight views that are used for relocalizing into another component.
    \item $W_{2}$ : Used to weight total observations ratio in the current run.
    \item $W_{3}$ : Used to weight total runs where the view was observed at least once. 
\end{itemize}

\subsection{View management algorithm}

Along with the data that helps to observe a view, our view data structure also stores some statistics that are used to determine its score. These include the total number of times the view has been observed in the current robot run, the time when the view was created, the number of runs the view was observed in, and whether the view has been used during relocalization into a component.

% View deletion algorithm
\begin{algorithm}[H]
    \caption{Find views for deletion}
    \label{lm_deletion_algorithm}
    \begin{algorithmic}[1]    	
    	\REQUIRE Set $\mathcal{V}$ of all the views
    	\IF {$|\mathcal{V}|$ $\le$ MIN\_VIEWS}
    	    \STATE return
    	\ENDIF
	    \STATE $\mathcal{V}_{keep} \leftarrow $ newlyCreatedViews()
	    \FOR {$v$ in $(\mathcal{V} \setminus \mathcal{V}_{keep})$}
	    	%\STATE $\mathit{score}\text{[}l\text{]} \leftarrow 0$
	    	%\STATE \{\texttt{n\_runs, n\_obs\_runs, n\_cur\_obs}\} $\leftarrow$ $\mathcal{H}(l)$
	    	%\STATE $\mathit{reloc} \leftarrow \text{importantForReLocalization($l$)}$
	    	\STATE% \# Score is based on the fraction of runs where lm was observed, fraction of current observations, and lm importance in re-localization \\
	    	{\em score}[$v$] $\leftarrow$
	    	\begin{equation*}
	    	\begin{aligned}
	    	\left( W_1 \times {\mathit{reloc}} + W_2 \times \frac{\mathit{n\_obs\_cur}}{\mathit{max\_obs}} + W_3 \times \frac{{\mathit{n\_obs\_runs}}}{\mathit{n\_runs}} \right)
	    	\end{aligned}
	    	\end{equation*}
	    	\IF {{\em score}[$v$] $>$ SCORE\_THRESHOLD}
	    		\STATE 
	    		$\mathcal{V}_{keep} \leftarrow \mathcal{V}_{keep} \cup \{v\}$
	    	\ENDIF
	    \ENDFOR
	    \STATE $\mathcal{V}_{delete} \leftarrow \mathcal{V} \setminus \mathcal{V}_{keep}$
	    \STATE sort($\mathcal{V}_{delete}$, {\em score}) \# Sort by score
	    %\STATE $\mathcal{V}_{delete}.\texttt{sort\_by\_score}()$
	    \STATE $D \leftarrow \varnothing$
	    \FOR {$v$ in $\mathcal{V}_{delete}$}
	    	\IF {numNearestNeighbors($v$, $D$) $<$ NN\_THRESHOLD}
	    		\STATE $\mathcal{V}_{keep} \leftarrow \mathcal{V}_{keep} \cup \{v\}$
	    	\ELSE
	    		\STATE $D \leftarrow D \cup \{v\}$
	    	\ENDIF
	    \ENDFOR
	    \STATE $\mathcal{V}_{delete} \leftarrow \mathcal{V} \setminus \mathcal{V}_{keep}$
	    \RETURN $\mathcal{V}_{delete}$    	
    \end{algorithmic}
\end{algorithm}

Let $\mathcal{V}_{keep}$ be the subset of views that should be kept and $\mathcal{V}$ be the set of all views. $\mathcal{V}_{keep}$ is initially empty, and is then populated with views that were created in the current run and observed at least once. Then a score is computed for the pre-existing views, based on the observation statistics from the current and previous runs, as well as the view's usefulness for relocalization. Views with scores higher than the threshold are also added to the set $\mathcal{V}_{keep}$. The view score is calculated as follows:
\begin{dmath*}
  W_{1} \times {\mathit{reloc}} + W_{2} \times \frac{\mathit{n\_obs\_cur}}{\mathit{max\_obs}} + W_{3} \times \frac{{\mathit{n\_obs\_runs}}}{\mathit{n\_runs}}
\end{dmath*}
where $n\_runs$ is the total number of runs when the view was present in the component, $n\_obs\_runs$ is the number of runs where the view was observed, $n\_obs\_cur$ is the number of observations of the view in the current run, and $max\_obs$ is the maximum number of observations of any view in the current run. $reloc$ is 1 if the view was used for relocalization and 0 otherwise. $W_{1}$, $W_{2}$, and $W_{3}$ are the weights.

The views that are currently in the set of $\mathcal{V}_{delete}$ containing $\mathcal{V} \setminus \mathcal{V}_{keep}$ are sorted by view score in ascending order. The number of nearby views for each view in $\mathcal{V}_{delete}$ is checked, and if it is less than the NN\_THRESHOLD, the view from $\mathcal{V}_{delete}$ and added to $\mathcal{V}_{keep}$. This constraint ensures a uniform spatial distribution of the views. In the end $\mathcal{V}_{delete}$ is the set of views that can be pruned without affecting the system's performance. The pseudo-code for the algorithm described above is presented in Algorithm~\ref{lm_deletion_algorithm}. Depending on the system, one might also want to add an upper bound on the total number of views per map, which can be determined experimentally based on the area of the space that is being mapped.

\section{Experiments and Results}

We have performed extensive experiments and analysis to evaluate the algorithm's performance and tune its parameters. We have collected run-time logs using a proprietary vacuum cleaning robot platform in several environments at different times of day with varied lighting conditions. The logs contain timestamped odometry readings, gyro readings, and camera images. Since most robot runs start from a dock / charging station, our logs start from the same location. We ran our visual SLAM system off-line on the logs, saving the map at the end of each run, so that it could be loaded for the next run. A saved map contains information about the SLAM graph, the views, and other supporting data. Trying to make an observation from a set of 2000 views would take more time than from a set of 200 views. Since this is dependent on the candidate view selection algorithm, CPU utilization results will vary based on what algorithm is being used. It follows that with increasing number of views, leading to an increase in the amount of time needed for making observations necessary for loop closures, the accuracy of the system will degrade. Because these are system dependent, and due to space constraints, we do not show any data related to this. Instead we concentrate on data related to the metrics that help us choose the ideal views for removal.

\subsection{Evaluation criteria}
In Section \ref{sec:method} we described the properties of the SLAM system that we have to preserve despite removing some of the views. The most important of those is the ability to relocalize in an existing map under various appearance changes of the environment. It is also important for the robot to be able to relocalize quickly, as opposed to wandering around for a long time looking for familiar views.

Thus we use the following criteria for measuring the robot's ability to relocalize:
\begin{itemize}
\item {\em Distance between cross-observations} is the distance the robot has to travel before observing a view created during a previous run.
\item {\em Fraction of cross-observed frames} is the fraction of the video frames in which the robot observed a view from a previous run.
\item {\em Relocalization distance} is the distance the robot has to travel before relocalizing, i. e. before observing the required number of views in another component (from a previous run in the same environment).
\end{itemize}
We compute these values for every robot run, and average them across all runs.\\

% \usepackage{booktabs}
\begin{table}[hb!]
\centering
\captionof{table}{View score weight selection ($W_2 = 1$, score threshold of 0.25 times max\_score) } \label{tab:lm_score_Weights_table_title}
\begin{tabular}{@{}ccccc@{}}
\toprule
\textbf{$W_1$} & \textbf{$W_3$} & \textbf{\begin{tabular}[c]{@{}c@{}}Avg. distance between\\ cross observations\end{tabular}} & \textbf{\begin{tabular}[c]{@{}c@{}}Fraction of cross\\ observed frames\end{tabular}} & \textbf{\begin{tabular}[c]{@{}c@{}}Growth \\ rate\end{tabular}} \\ \midrule
0           & 0           & 1.429                                                                                       & 0.054                                                                                & 0.7                                                             \\
0           & 0.5         & 0.606                                                                                       & 0.106                                                                                & 14.45                                                           \\
0           & 1           & 0.604                                                                                       & 0.107                                                                                & 15.2                                                            \\
0           & 1.5         & 0.607                                                                                       & 0.106                                                                                & 14.4                                                            \\
0           & 2           & 0.572                                                                                       & 0.107                                                                                & 14.65                                                           \\
0           & 2.5         & 0.602                                                                                       & 0.107                                                                                & 14.2                                                            \\
\textbf{0}           & \textbf{3}           & \textbf{0.727}                                                                                       & \textbf{0.081}                                                                                & \textbf{4.1}                                                             \\ \midrule
0.5         & 0           & 1.093                                                                                       & 0.057                                                                                & 1.15                                                            \\
0.5         & 0.5         & 0.61                                                                                        & 0.104                                                                                & 15.05                                                           \\
0.5         & 1           & 0.598                                                                                       & 0.109                                                                                & 14.45                                                           \\
0.5         & 1.5         & 0.618                                                                                       & 0.105                                                                                & 14.85                                                           \\
0.5         & 2           & 0.603                                                                                       & 0.106                                                                                & 14.7                                                            \\
\textbf{0.5}         & \textbf{2.5}         & \textbf{0.699}                                                                                       & \textbf{0.085}                                                                                & \textbf{3.45}                                                            \\
\textbf{0.5}         & \textbf{3}           & \textbf{0.748}                                                                                       & \textbf{0.086}                                                                                & \textbf{3.85}                                                            \\ \midrule
1           & 0           & 1.196                                                                                       & 0.058                                                                                & 1                                                               \\
1           & 0.5         & 0.673                                                                                       & 0.104                                                                                & 14.45                                                           \\
1           & 1           & 0.611                                                                                       & 0.104                                                                                & 14.35                                                           \\
1           & 1.5         & 0.615                                                                                       & 0.104                                                                                & 15.05                                                           \\
1           & 2           & 0.841                                                                                       & 0.077                                                                                & 2.45                                                            \\
\textbf{1}           & \textbf{2.5}         & \textbf{0.794}                                                                                       & \textbf{0.082}                                                                                & \textbf{3.65}                                                            \\
\textbf{1}           & \textbf{3}           & \textbf{0.673}                                                                                       & \textbf{0.086}                                                                                & \textbf{4.15}                                                            \\ \midrule
1.5         & 0           & 1.095                                                                                       & 0.061                                                                                & 1.2                                                             \\
1.5         & 0.5         & 0.565                                                                                       & 0.108                                                                                & 14.35                                                           \\
1.5         & 1           & 0.603                                                                                       & 0.105                                                                                & 14.5                                                            \\
1.5         & 1.5         & 0.957                                                                                       & 0.067                                                                                & 1.3                                                             \\
1.5         & 2           & 0.914                                                                                       & 0.073                                                                                & 2                                                               \\
\textbf{1.5}         & \textbf{2.5}         & \textbf{0.781}                                                                                       & \textbf{0.082}                                                                                & \textbf{3.3}                                                             \\
\textbf{1.5}         & \textbf{3}           & \textbf{0.711}                                                                                       & \textbf{0.084}                                                                                & \textbf{4.25}                                                            \\ \midrule
2           & 0           & 1.183                                                                                       & 0.059                                                                                & 0.95                                                            \\
2           & 0.5         & 0.67                                                                                        & 0.105                                                                                & 15.1                                                            \\
2           & 1           & 1.211                                                                                       & 0.059                                                                                & 1.15                                                            \\
2           & 1.5         & 0.942                                                                                       & 0.068                                                                                & 1.55                                                            \\
2           & 2           & 0.844                                                                                       & 0.075                                                                                & 2.3                                                             \\
\textbf{2}           & \textbf{2.5}         & \textbf{0.689}                                                                                       & \textbf{0.086}                                                                                & \textbf{3.45}                                                            \\
\textbf{2}           & \textbf{3}           & \textbf{0.774}                                                                                       & \textbf{0.085}                                                                                & \textbf{4.25}                                                            \\ \bottomrule
\end{tabular}
\end{table}

The other goal of our algorithm is to limit the growth of the number of views in the graph. To measure that, we define the {\em view growth rate} $G$ as follows:
\begin{dmath*}
G = \frac{v_n - v_2}{n-1}
\end{dmath*}
where $v_i$ is the number of views at the end of the $i$-th run, and $n$ is the total number of runs. We start from the second run because in the first run the robot creates a brand new map, rather than relocalizing into an existing map.

%The rest of the experiments section is divided into three parts. In section \ref{sec:lm_score_weights}, we describe our methodology for choosing the appropriate weights for calculating the view score. In section \ref{sec:nn_threshold}, we present the selection of the various thresholds for the nearest neighbor check in the algorithm to ensure a uniform spatial distribution of the views in the map. Finally, in section \ref{sec:lifelong_mapping}, we show lifelong mapping results from different environments with many runs ( $>$ 100 ).

\subsection{View score weights selection}
\label{sec:lm_score_weights}
% Plot the graph of average view score percentage, and then see where the cluster is formed ...
% Based on that, we can choose an appropriate threshold

We divide the algorithm parameters into two subsets: the weights for computing the view scores, and the parameters for counting the nearest neighbor views.  It is not practical to tune both subsets simultaneously,  because the scoring function and the nearest neighbor check work against each other. Therefore, first, we disable the nearest neighbor check and tune the view score weights using the distance between cross-observations, the fraction of cross-observed frames, and the view growth rate.

We want to choose the weights such that the distance between cross-observations is low,  the fraction of cross-observed frames is high, and the growth rate is low. The values in Table~\ref{tab:lm_score_Weights_table_title} were generated by running the robot logs of environment A sequentially 22 times.  We used a threshold of 5.0 for the growth rate, 0.8m for the distance between-cross observations, and 0.075 for the fraction of cross-observed frames. From the table we see that the weights of (0, 1, 3), (0.5, 1, 2.5), (0.5, 1, 3), (1, 1, 2.5), (1, 1, 3), (1.5, 1, 2.5), (1.5, 1, 3), (2, 1, 2.5), and (2, 1, 3) are good choices.

\begin{figure}[h!]
    \centering
    \includegraphics[width=0.35\textwidth]{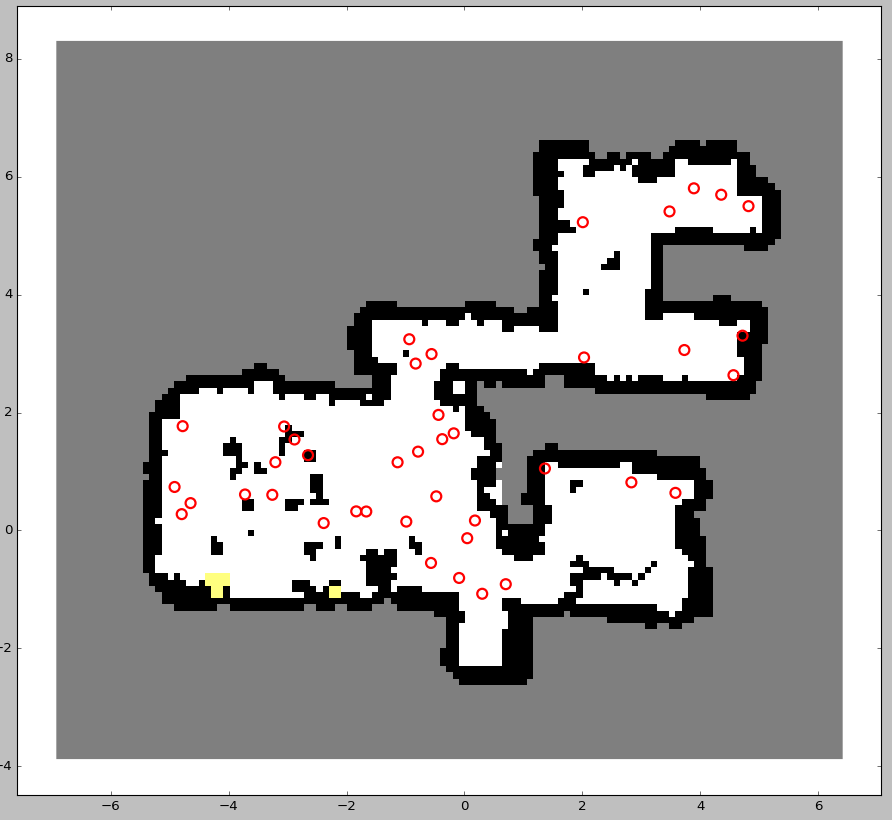}
    \caption{Uneven distribution of views after 20 runs in an environment with only view score based pruning.}
    \label{fig:sparse_landmarks}
\end{figure}

\begin{figure}[h!]
    \centering
    \includegraphics[width=0.35\textwidth]{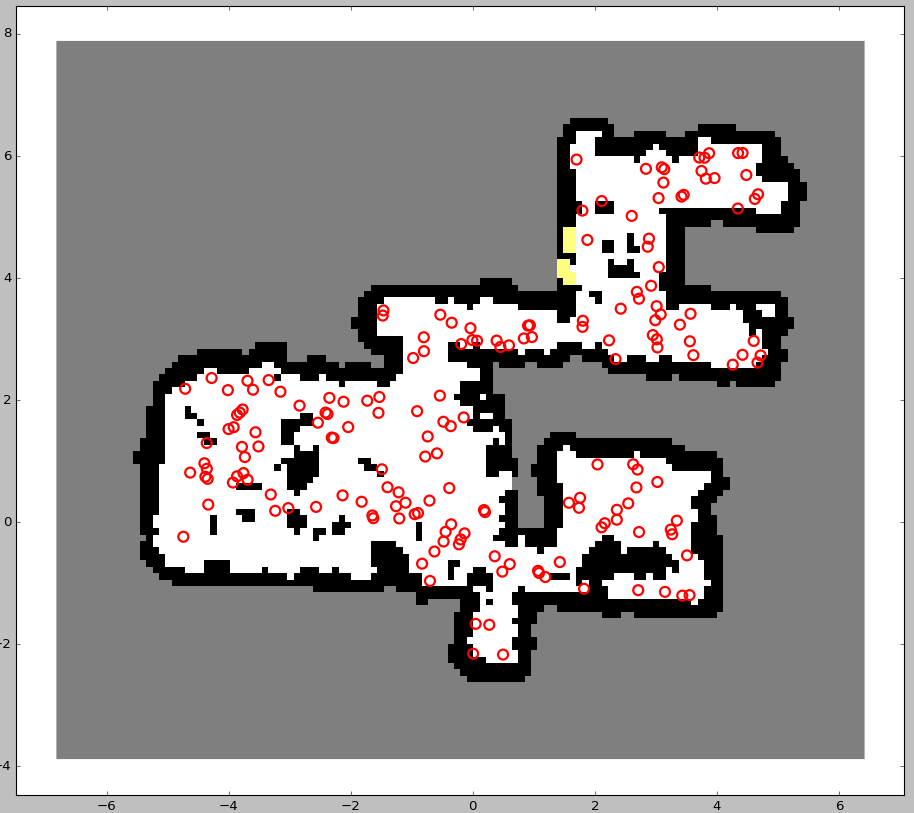}
    \caption{Uniform distribution of views after 20 runs in the same environment as the previous figure after applying the nearest neighbor constraint.}
    \label{fig:uniform_landmarks}
\end{figure}

\subsection{Nearest neighbor threshold selection}
\label{sec:nn_threshold}
The nearest neighbor function returns the number of neighboring views present in a voxel of size specified by NN\_VOXEL\_SIZE surrounding the view that is being queried. After views are marked for deletion based on the view score, the nearest neighbor check is performed. The nearest neighbor constraint is needed to ensure a good spatial distribution (see Fig.~\ref{fig:uniform_landmarks}) of views across the entire map. Without this additional constraint, maps can end up with an uneven spatial distribution of views (see Fig.~\ref{fig:sparse_landmarks}).

A small voxel size will mean that a view might end up having too few neighbors, and a high nearest neighbor threshold will mean too many minimum neighboring views. In either case, the nearest neighbor constraint would restrict pruning. A uniform spatial distribution would be maintained but the growth rate of views would be high. Whereas a very large voxel size will mean too many neighbors, and a small nearest neighbor threshold will restrict the number of views. In these scenarios, the nearest neighbor constraint wouldn't ensure a uniform spatial distribution but would restrict the growth rate. So, the choice of the nearest neighbor voxel parameters and the nearest neighbor threshold needs to be carefully chosen so as to have a good uniform distribution and a reasonable growth rate.

\begin{figure*}[t]
    \centering
        \begin{subfigure}[b]{0.34\textwidth}
            \includegraphics[width=\textwidth]{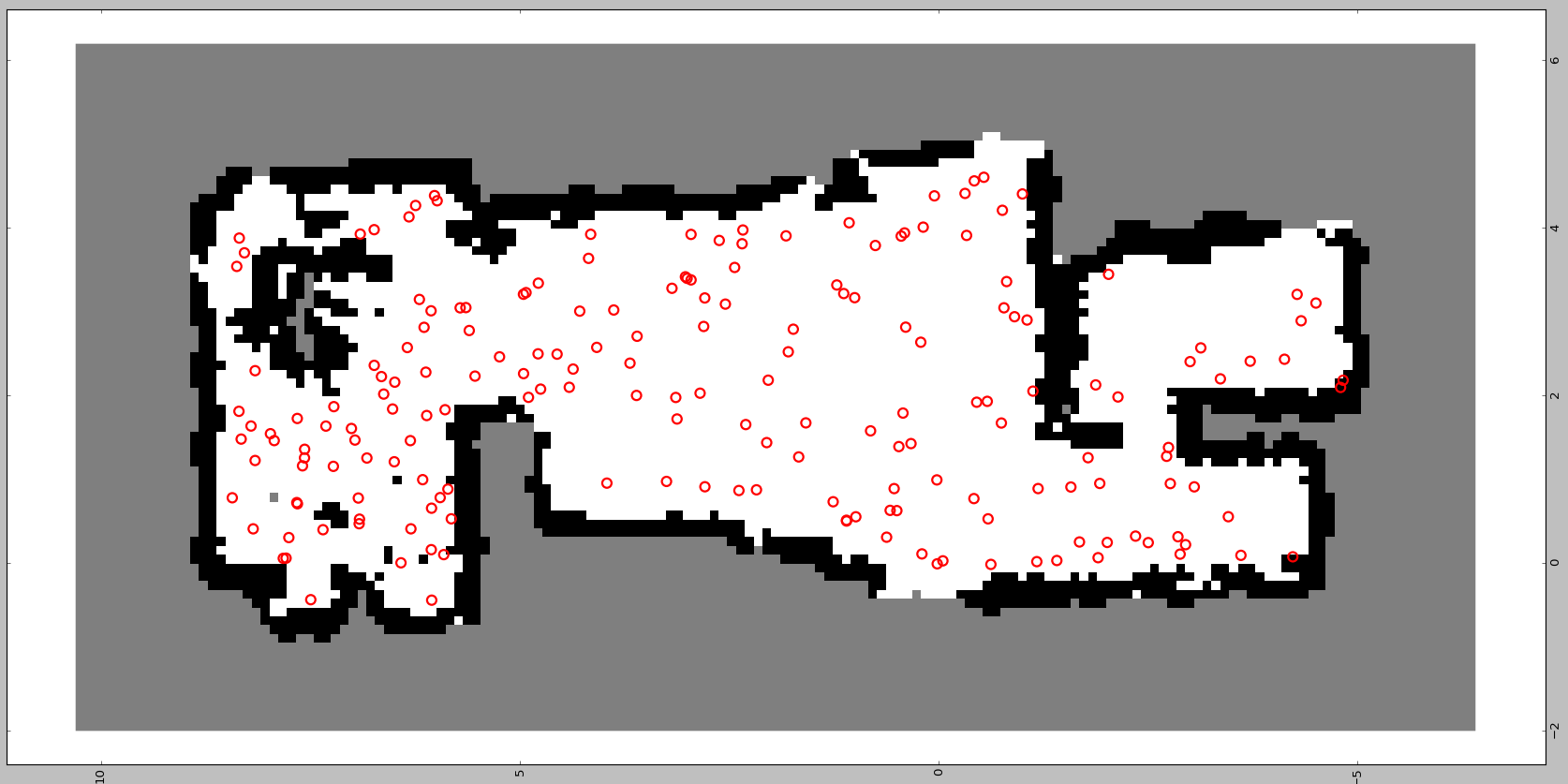}
            \label{fig:map_manju_house}
        \end{subfigure}%
        ~
        \begin{subfigure}[b]{0.18\textwidth}
            \includegraphics[width=\textwidth]{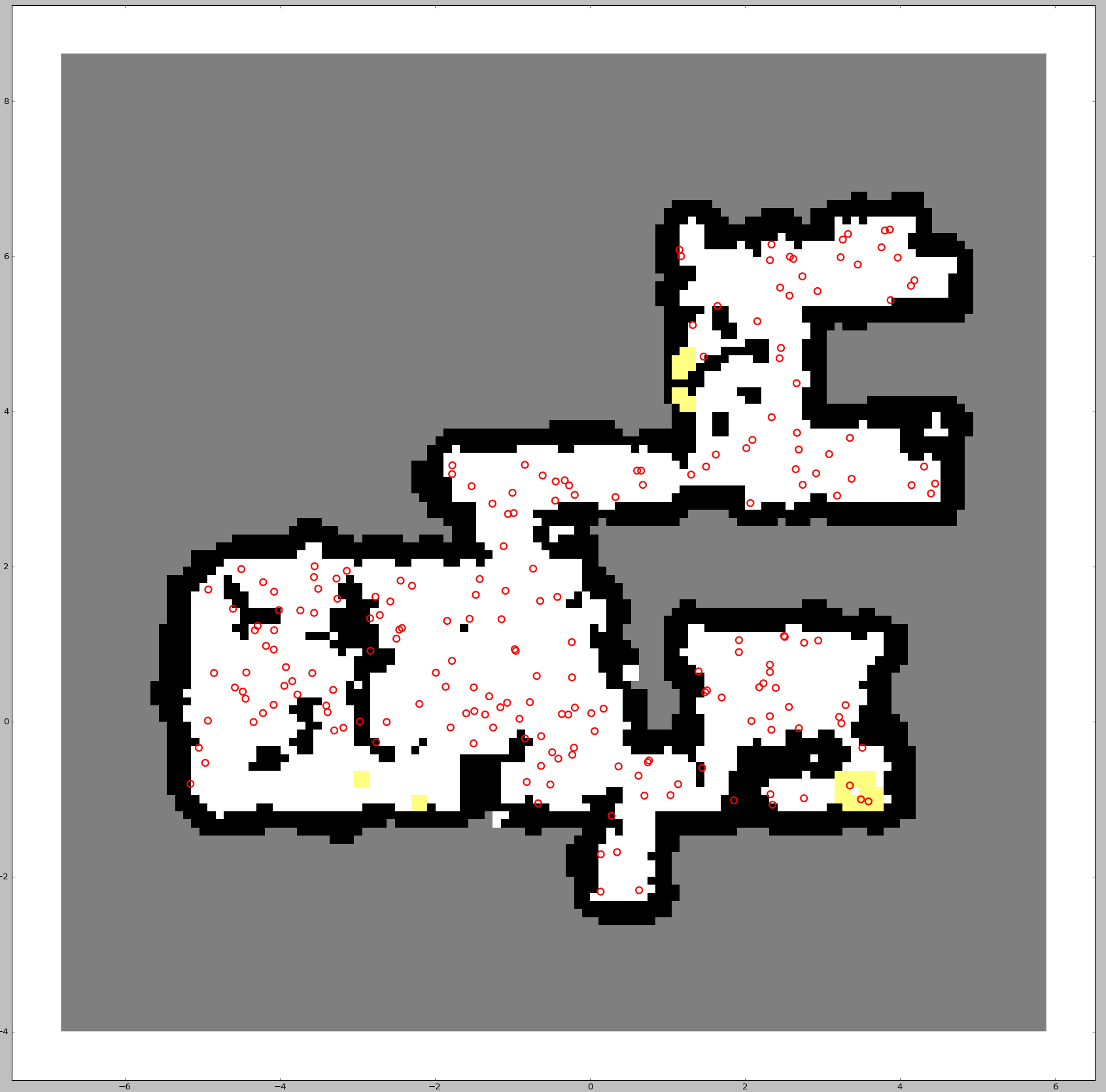}
            \label{fig:map_nandan_house}
        \end{subfigure}
        \begin{subfigure}[b]{0.18\textwidth}
            \includegraphics[width=\textwidth]{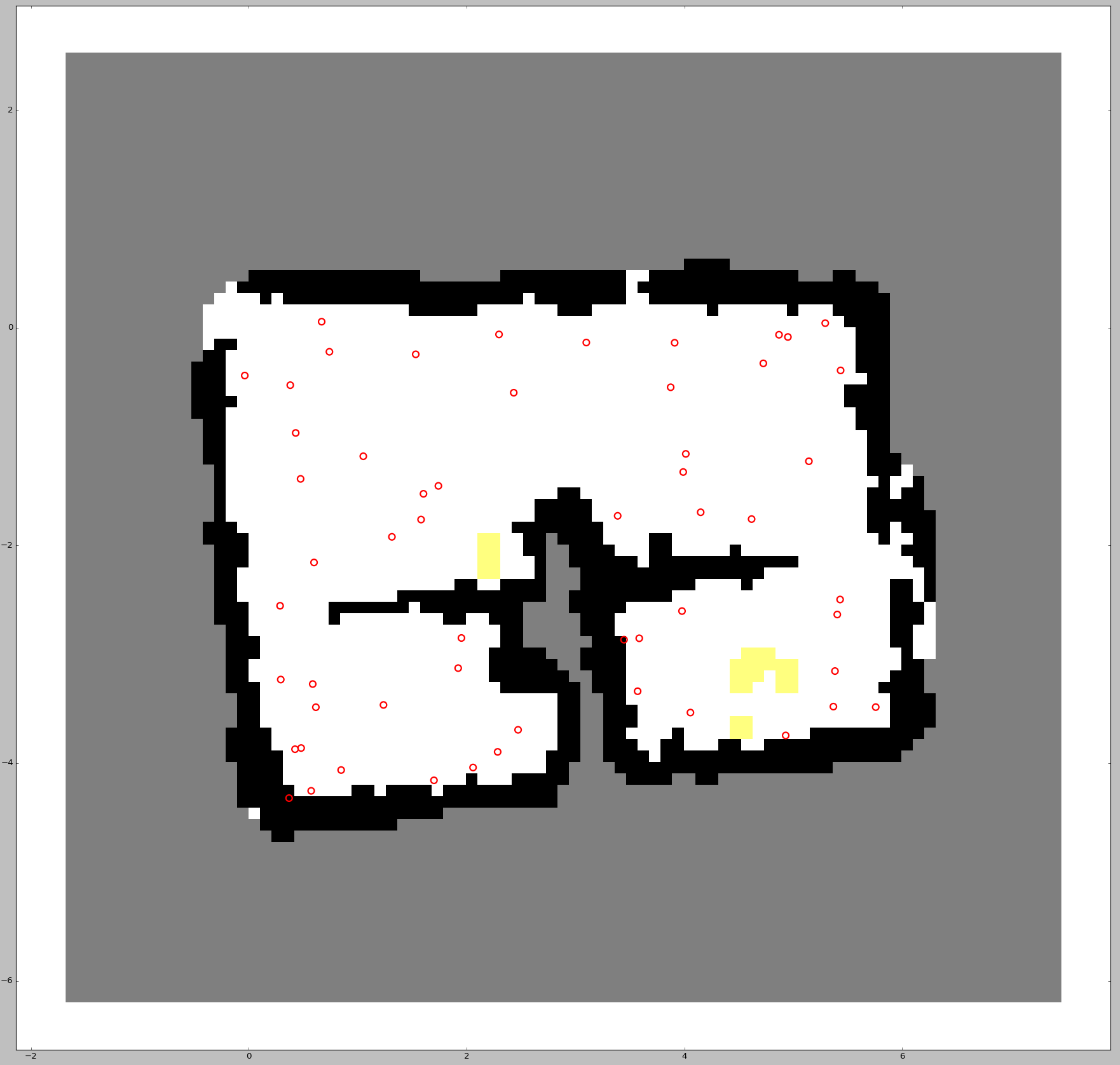}
            \label{fig:map_pasadena_test_area_1}
        \end{subfigure}
        ~
        \begin{subfigure}[b]{0.16\textwidth}
            \includegraphics[width=\textwidth]{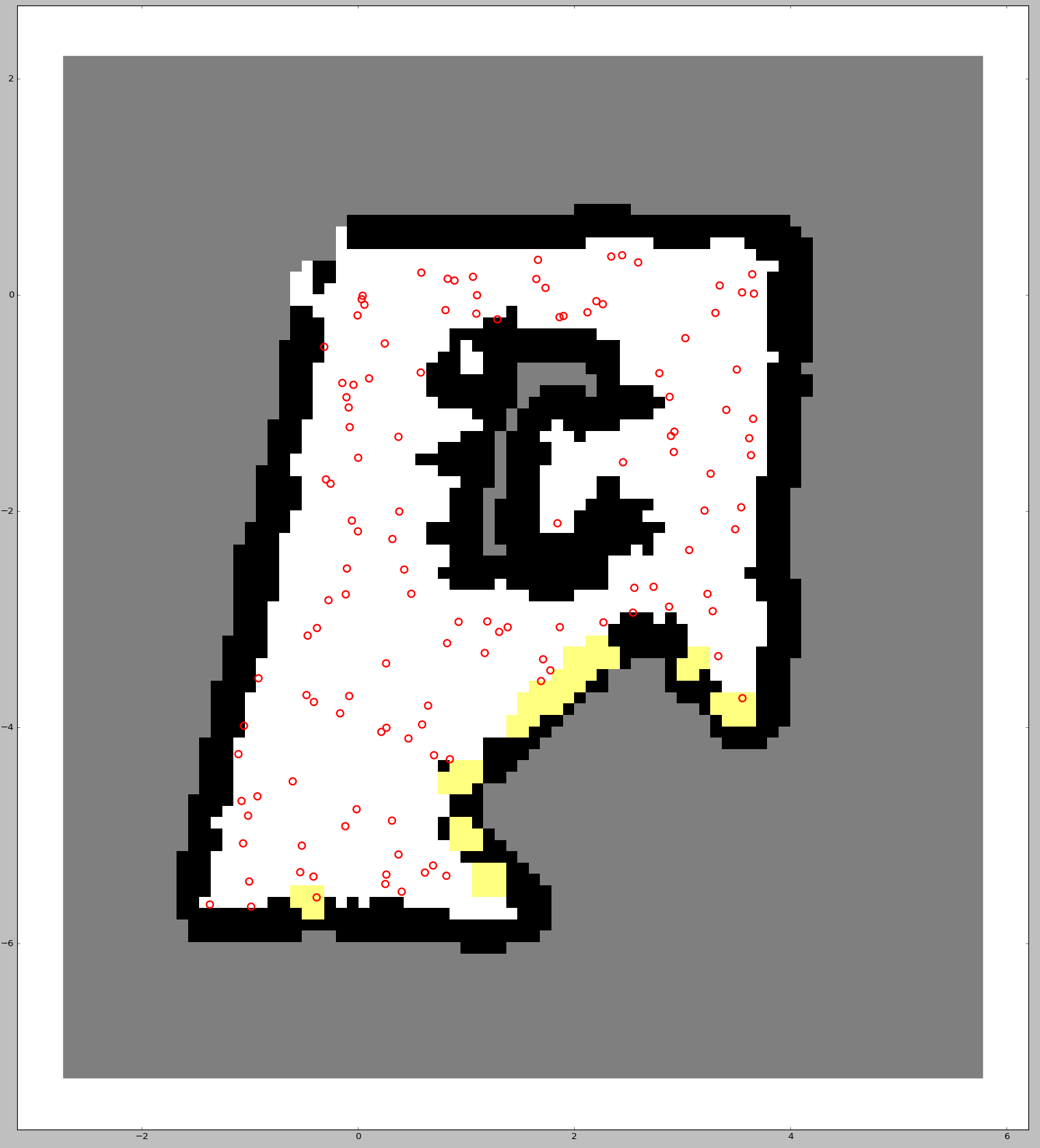}
            \label{fig:map_pasadena_test_area_2}
        \end{subfigure}
    \caption{Occupancy maps with overlaid views of different environments A, B, C, and D (from left to right). Env A and Env B are large with around 650 and 500 $ft^2$ of navigable area respectively. Env C and Env D have about 300 $ft^2$ of navigable area. Logs from Env A and Env B have more texture on walls and furniture than logs from Env C and Env D. All logs were collected with varied lighting conditions, both during daytime and nighttime.}
    \label{fig:maps_with_views}
\end{figure*}

\begin{figure*}[ht]
    \centering
        \begin{subfigure}[b]{0.4\textwidth}
            \includegraphics[width=\textwidth]{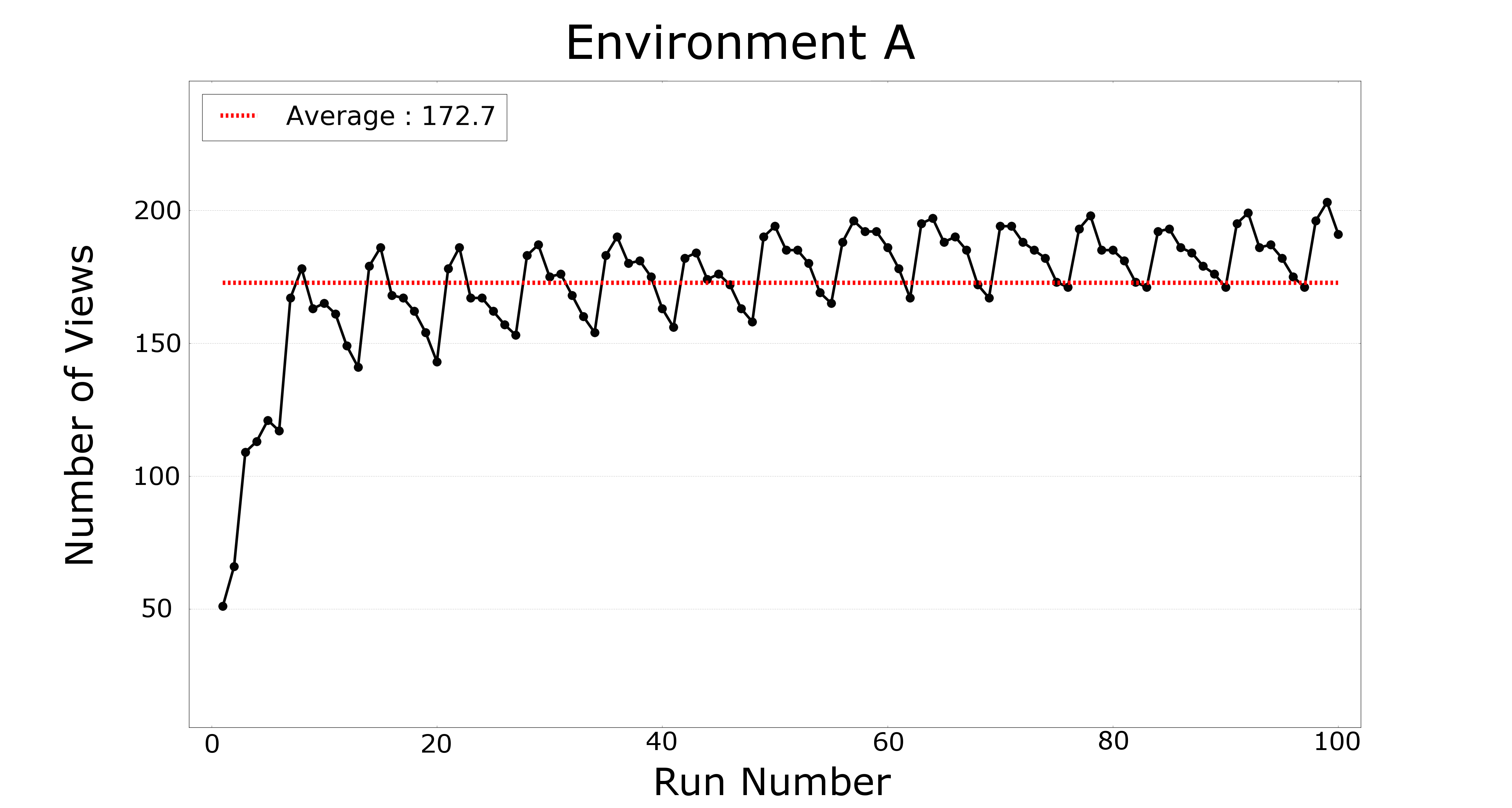}
            \label{fig:growth_manju_house}
        \end{subfigure}
        ~
        \begin{subfigure}[b]{0.4\textwidth}
            \includegraphics[width=\textwidth]{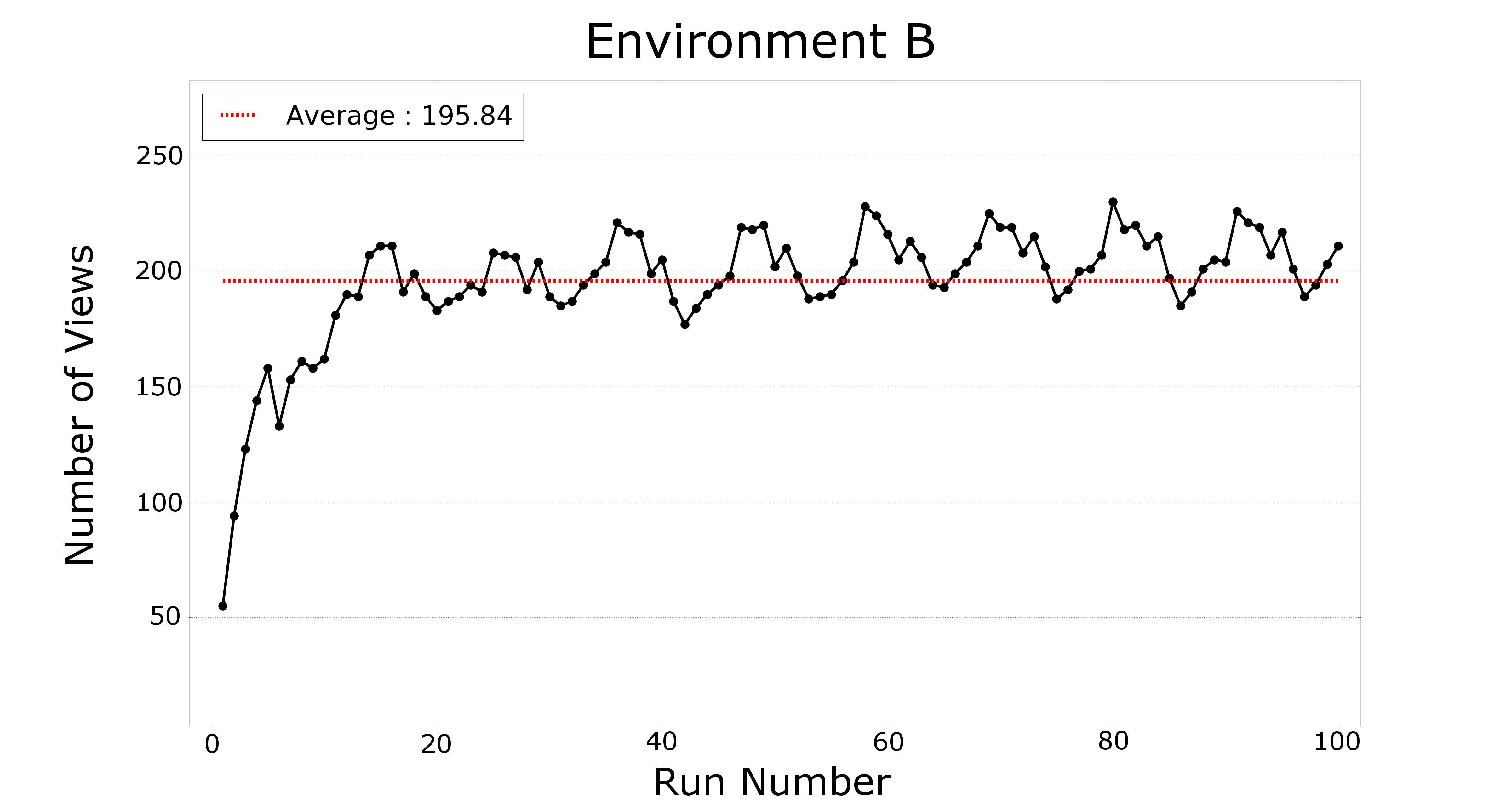}
            \label{fig:growth_nandan_house}
        \end{subfigure}
        ~
        \begin{subfigure}[b]{0.4\textwidth}
            \includegraphics[width=\textwidth]{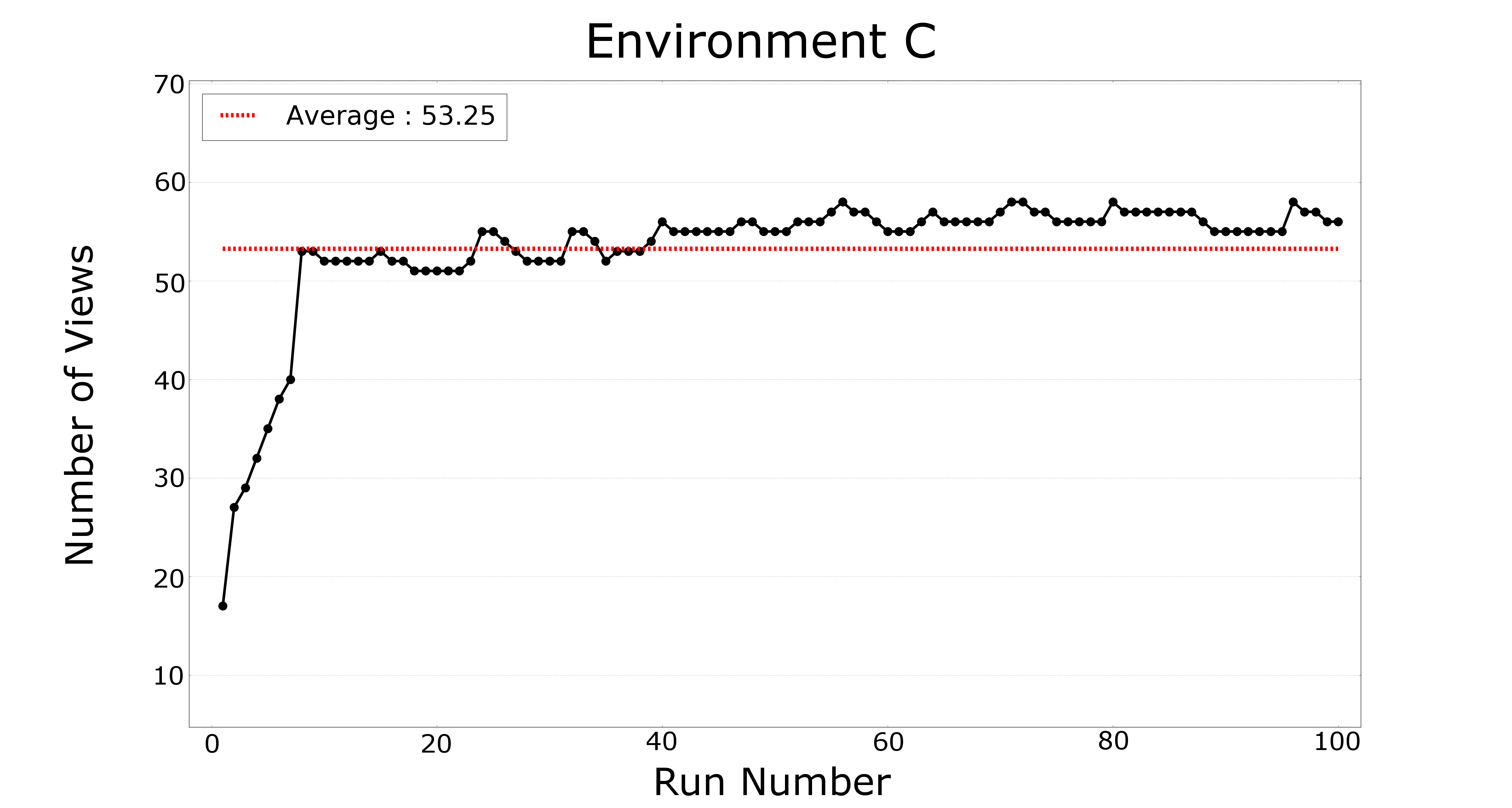}
            \label{fig:growth_pasadena_test_area_1}
        \end{subfigure}
        ~
        \begin{subfigure}[b]{0.4\textwidth}
            \includegraphics[width=\textwidth]{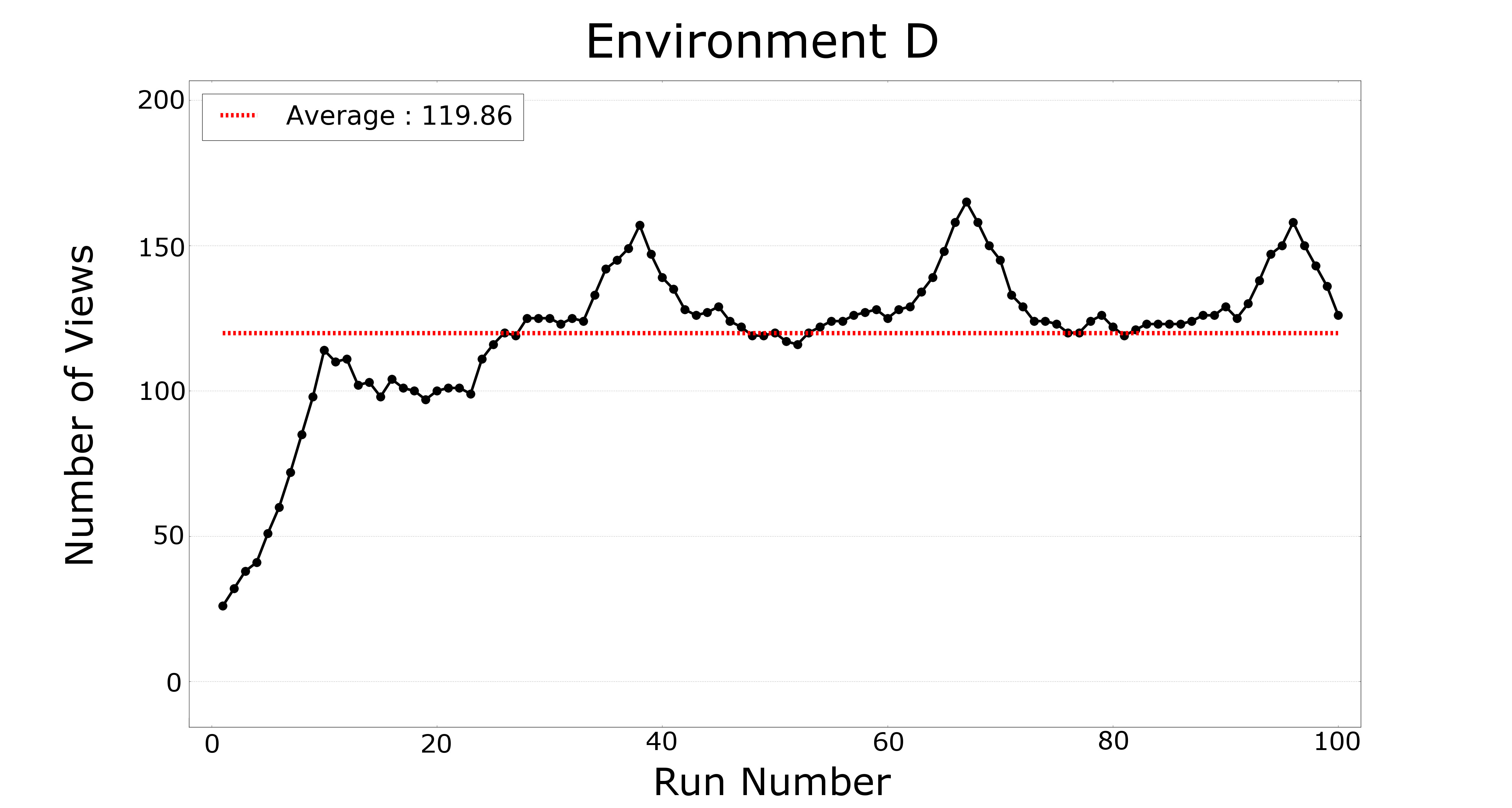}
            \label{fig:growth_pasadena_test_area_2}
        \end{subfigure}
    \caption{The growth of views over 100 runs in the four environments from Fig.~\ref{fig:maps_with_views}. The total number of views stabilize after 5 to 10 runs.}
    \label{fig:growth_all_envs}
\end{figure*}

\begin{figure*}[ht]
    \centering
        \begin{subfigure}[b]{0.4\textwidth}
            \includegraphics[width=\textwidth]{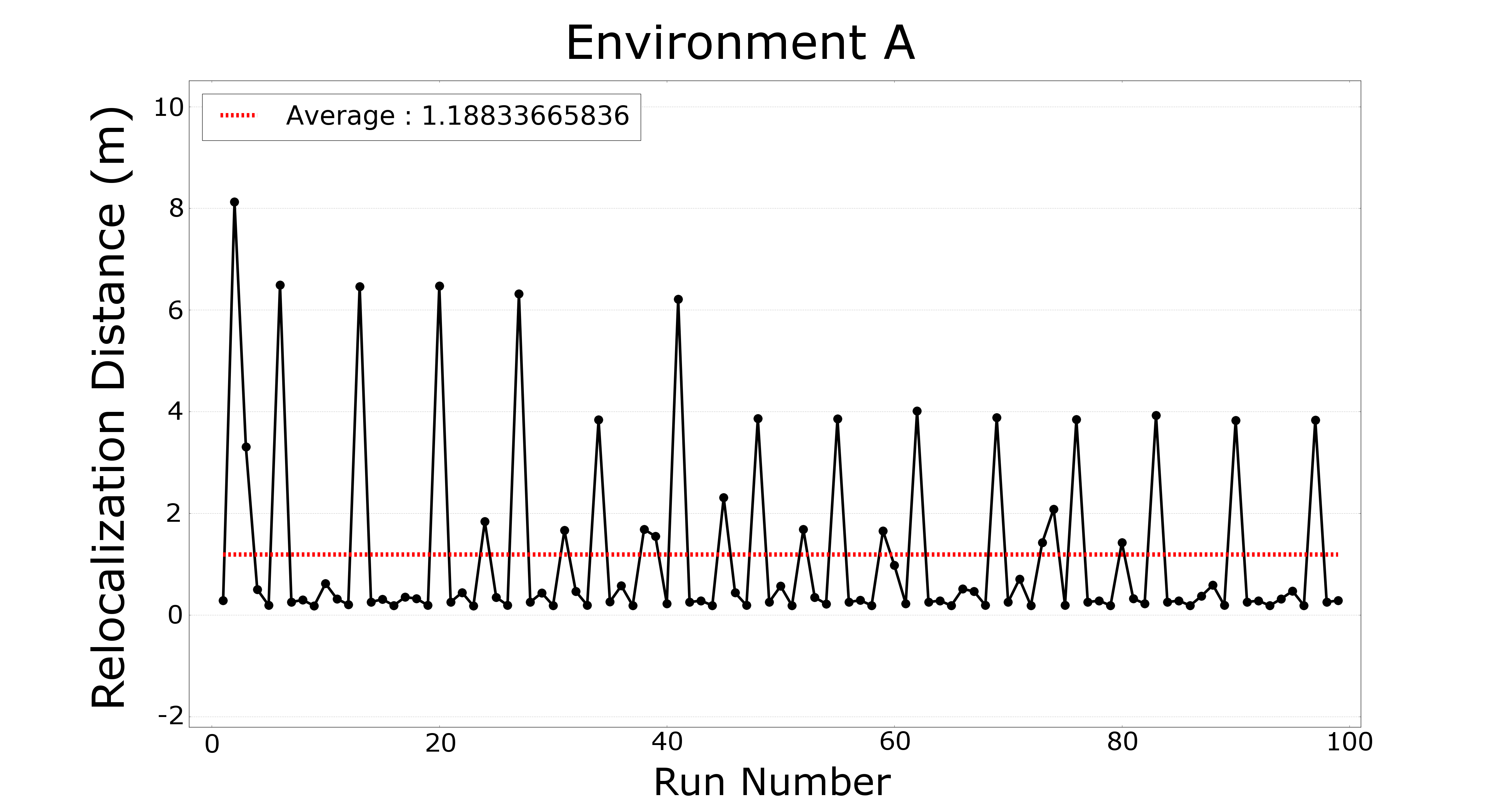}
            \label{fig:reloc_distance_manju_house}
        \end{subfigure}
        ~
        \begin{subfigure}[b]{0.4\textwidth}
            \includegraphics[width=\textwidth]{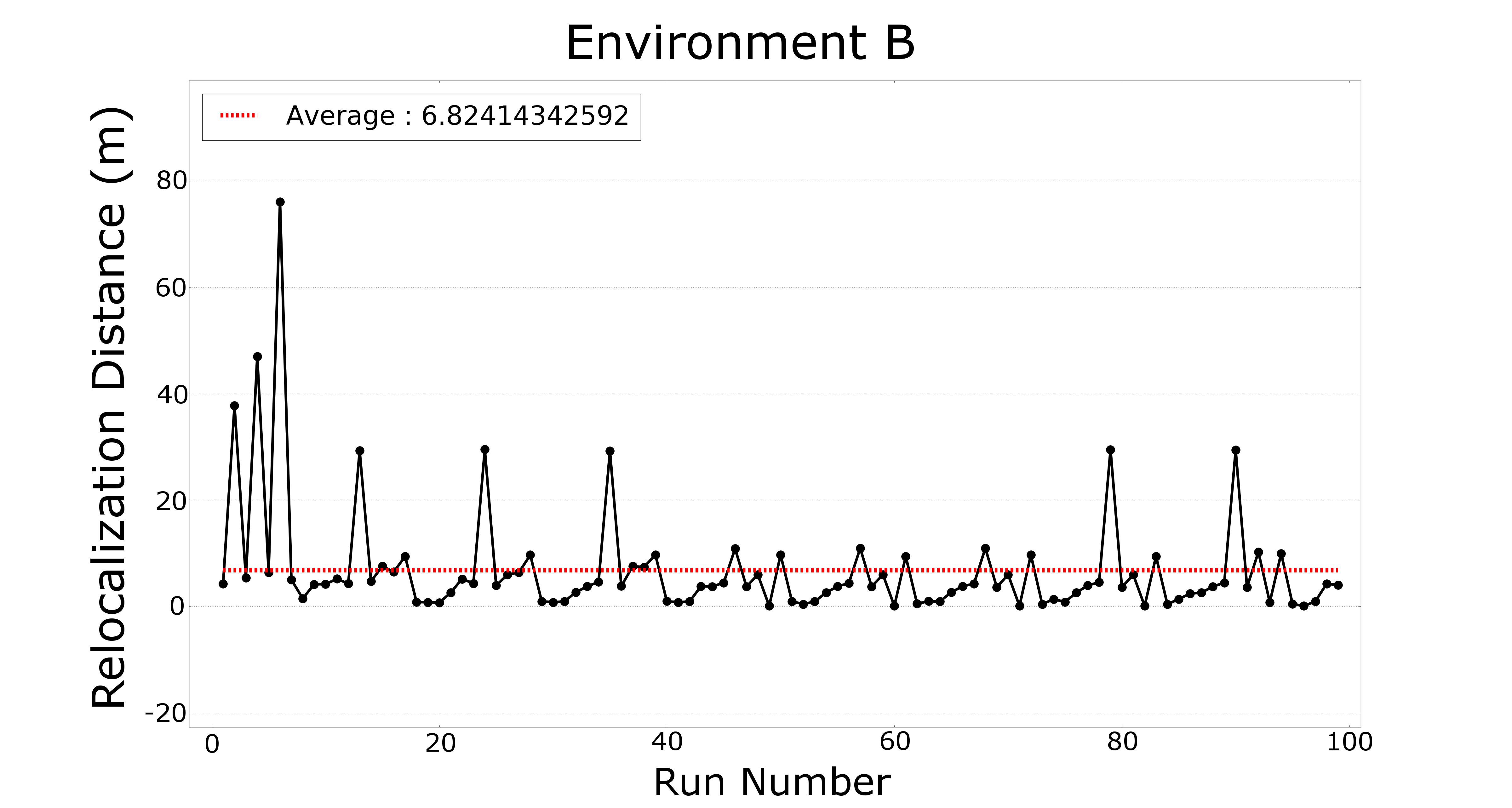}
            \label{fig:reloc_distance_nandan_house}
        \end{subfigure}
        ~
        \begin{subfigure}[b]{0.4\textwidth}
            \includegraphics[width=\textwidth]{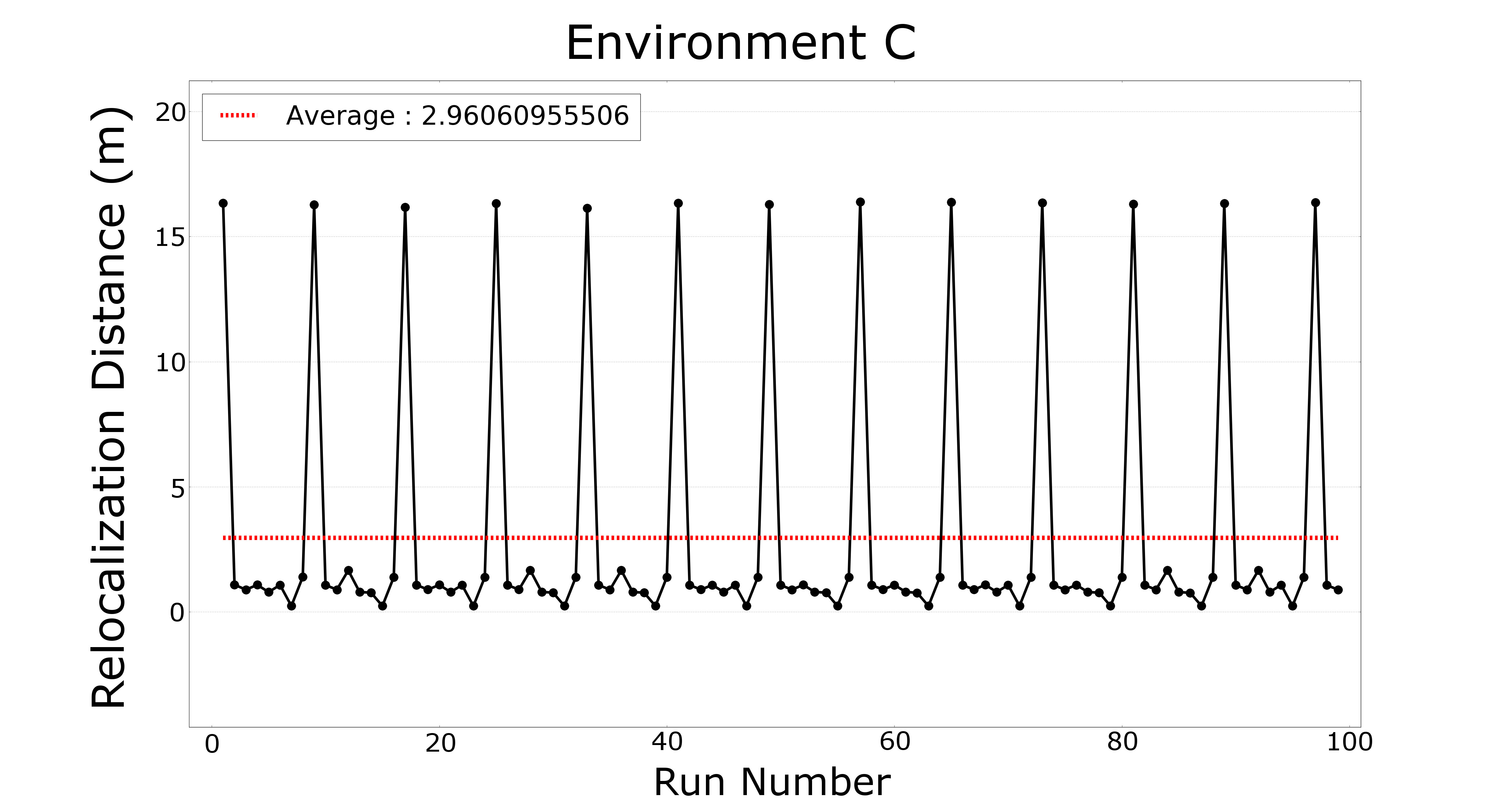}
            \label{fig:reloc_distance_pasadena_test_area_1}
        \end{subfigure}
        ~
        \begin{subfigure}[b]{0.4\textwidth}
            \includegraphics[width=\textwidth]{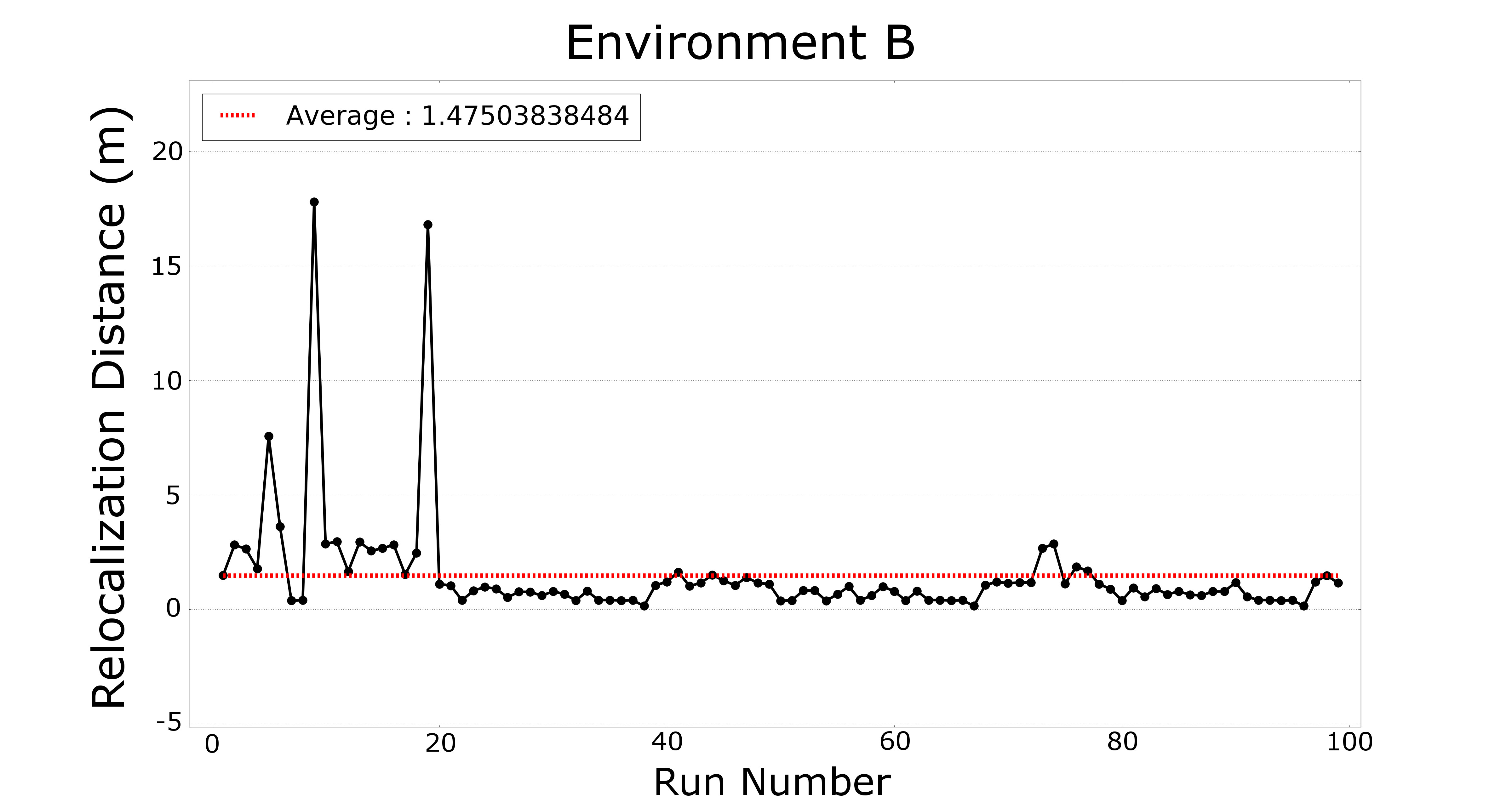}
            \label{fig:reloc_distance_pasadena_test_area_2}
        \end{subfigure}
    \caption{Plot of the relocalization distance, the lower the better. The relocalization distance is usually low in most cases, and in some maps there are large spikes in the beginning which is due to the system taking time to observe views from a loaded map due to very different lighting conditions.}
    \label{fig:reloc_distance_all_envs}
\end{figure*}

\begin{figure*}[ht]
    \centering
        \begin{subfigure}[b]{0.4\textwidth}
            \includegraphics[width=\textwidth]{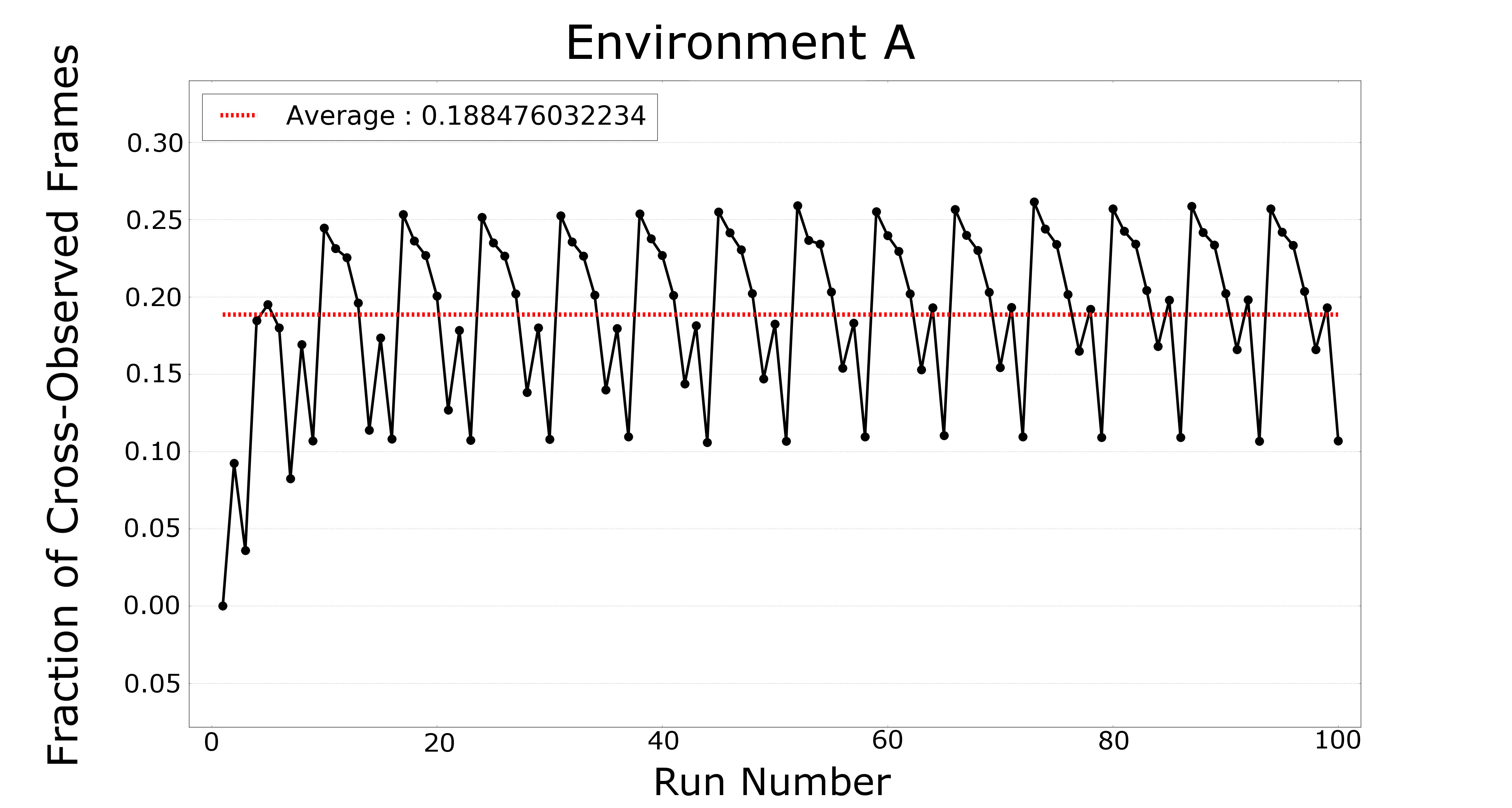}
            \label{fig:frac_x_observed_manju_house}
        \end{subfigure}
        ~
        \begin{subfigure}[b]{0.4\textwidth}
            \includegraphics[width=\textwidth]{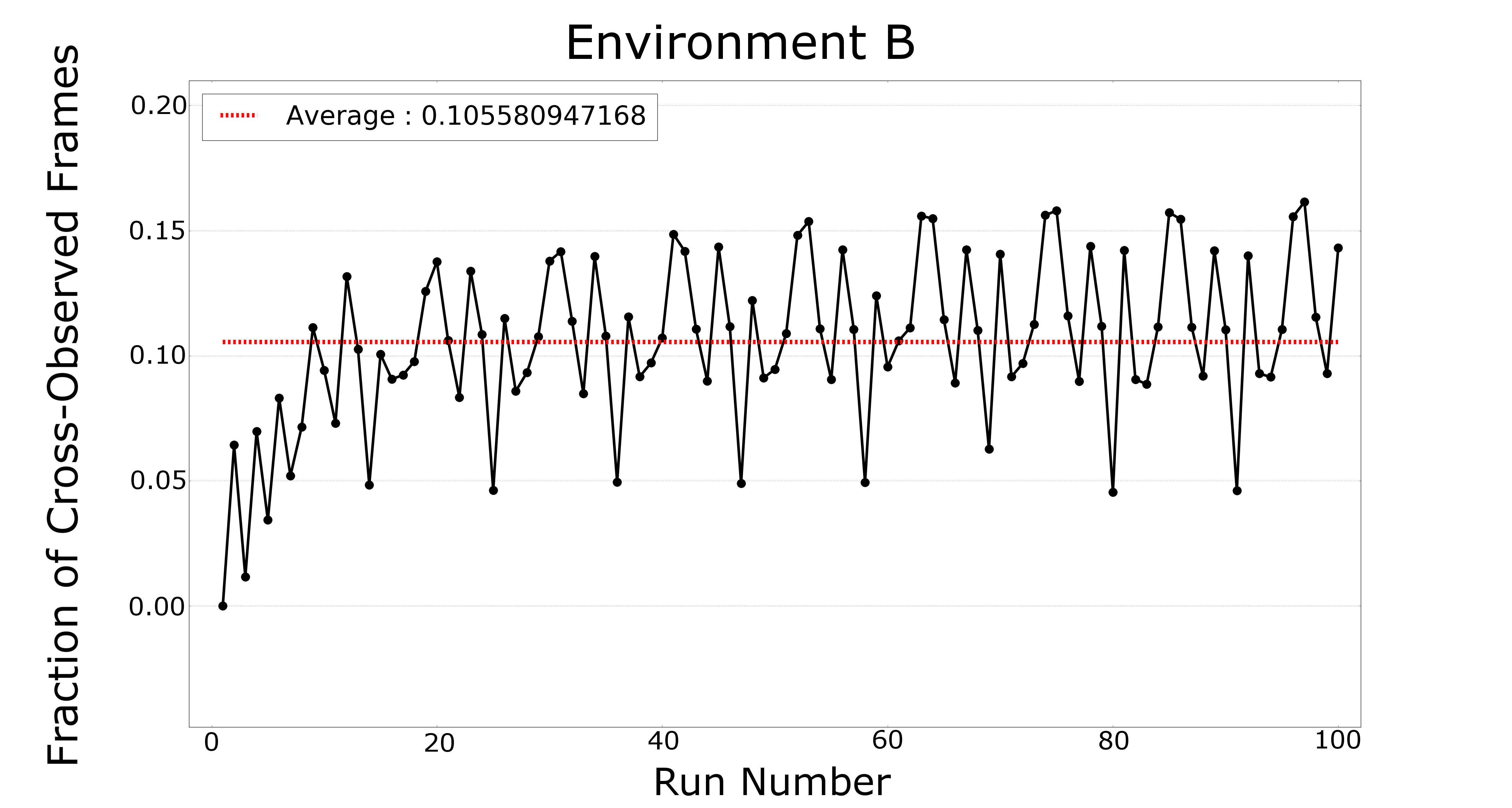}
            \label{fig:frac_x_observed_nandan_house}
        \end{subfigure}
        ~
        \begin{subfigure}[b]{0.4\textwidth}
            \includegraphics[width=\textwidth]{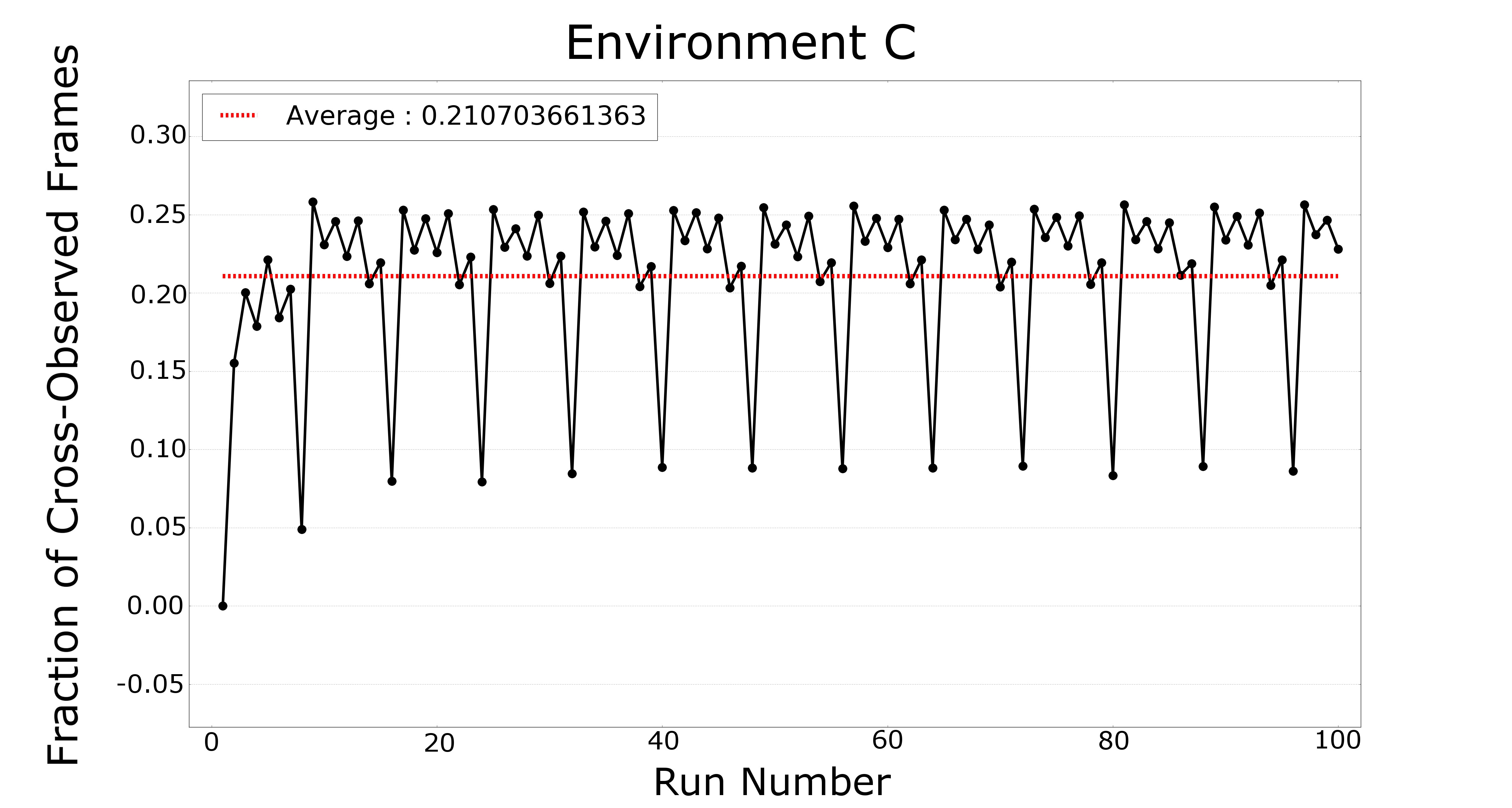}
            \label{fig:frac_x_observed_pasadena_test_area_1}
        \end{subfigure}
        ~
        \begin{subfigure}[b]{0.4\textwidth}
            \includegraphics[width=\textwidth]{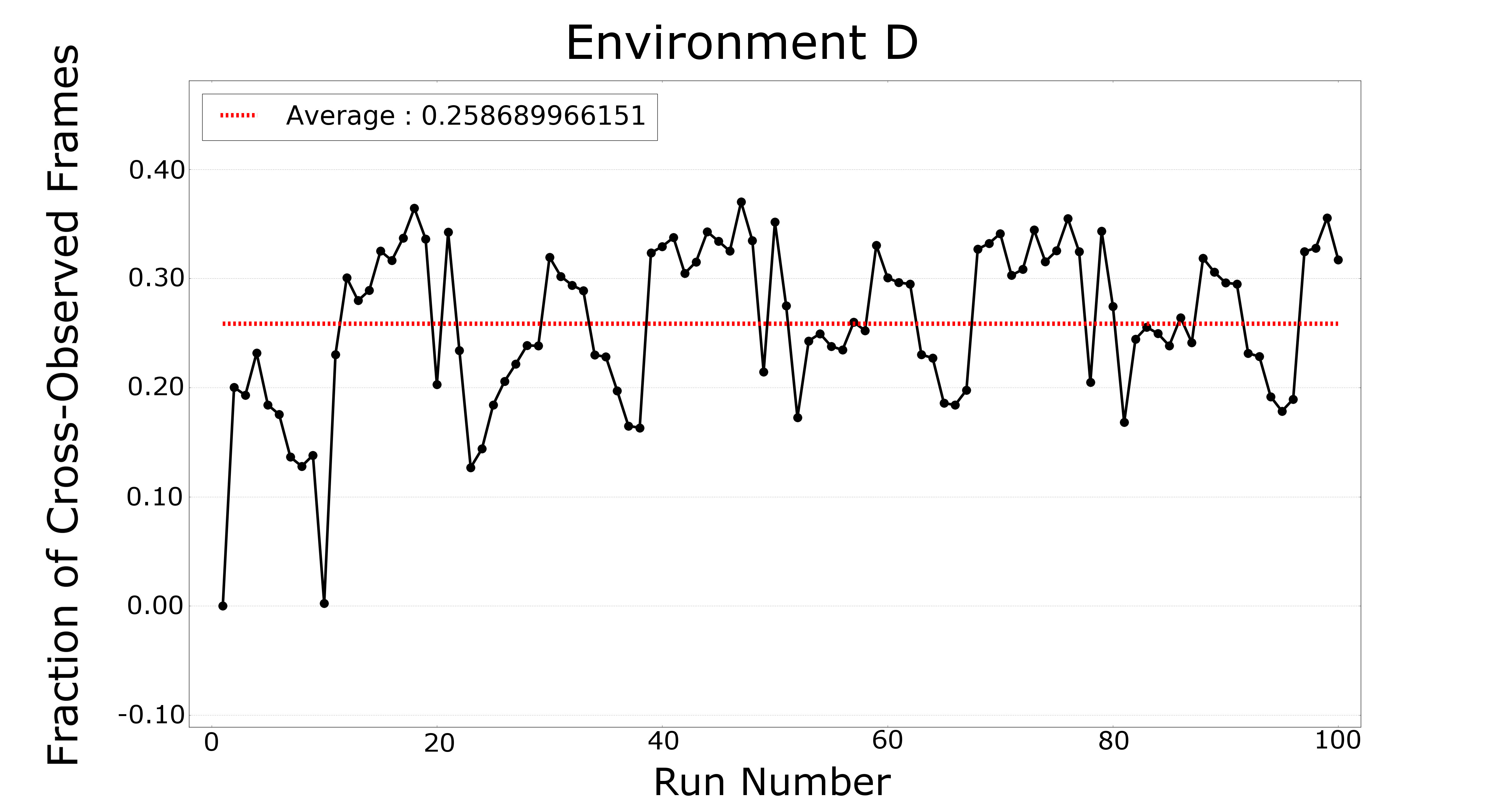}
            \label{fig:frac_x_observed_pasadena_test_area_2}
        \end{subfigure}
    \caption{Plot of the fraction of cross observed frames, the higher the better. This plot gives an idea of how the fraction of cross observations change over multiple runs with the view management system in place.}
    \label{fig:frac_x_observed_all_envs}
\end{figure*}

\begin{figure*}[ht]
    \centering
        \begin{subfigure}[b]{0.38\textwidth}
            \includegraphics[width=\textwidth]{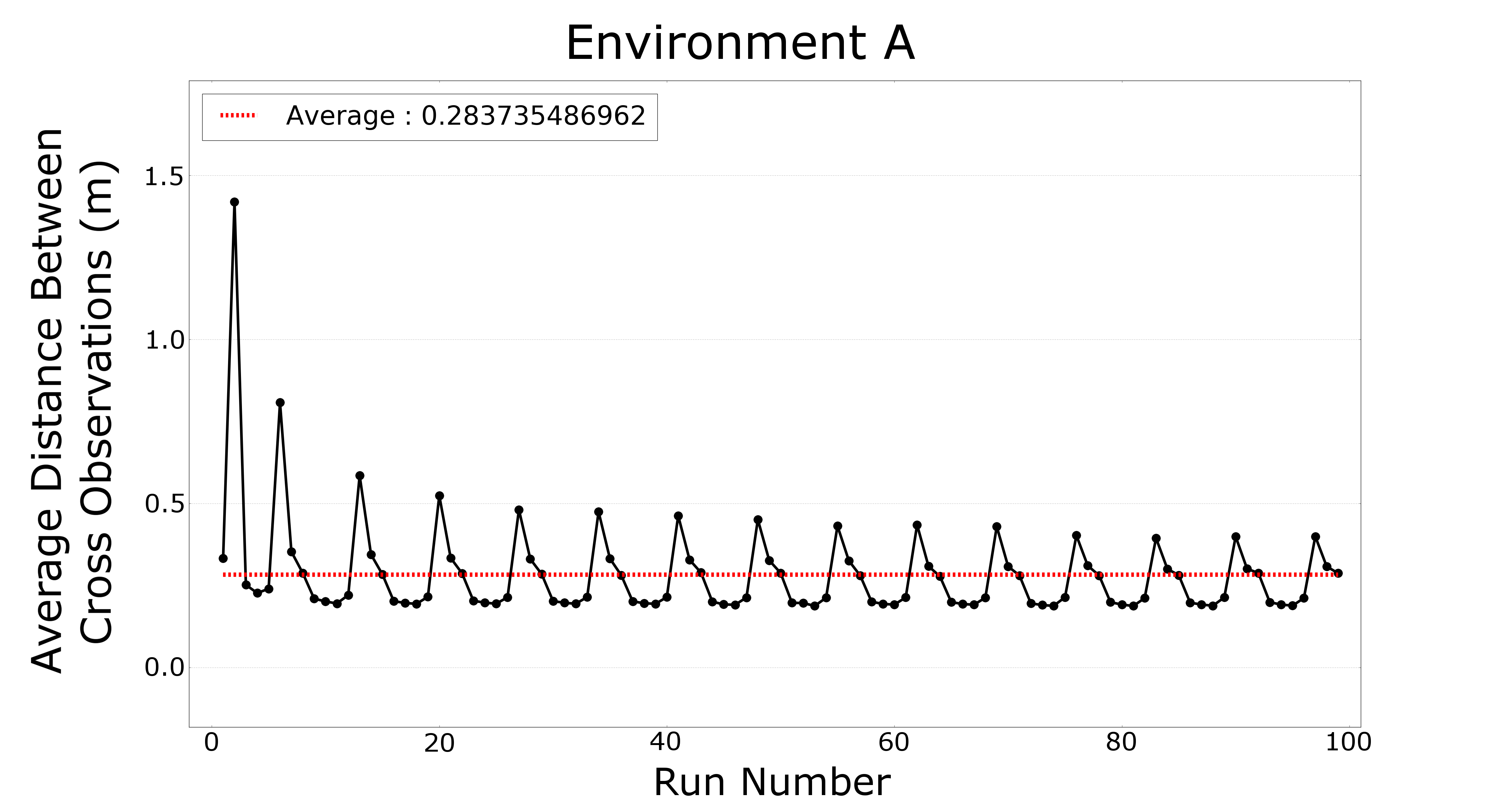}
            \label{fig:avg_dist_x_obs_manju_house}
        \end{subfigure}
        ~
        \begin{subfigure}[b]{0.38\textwidth}
            \includegraphics[width=\textwidth]{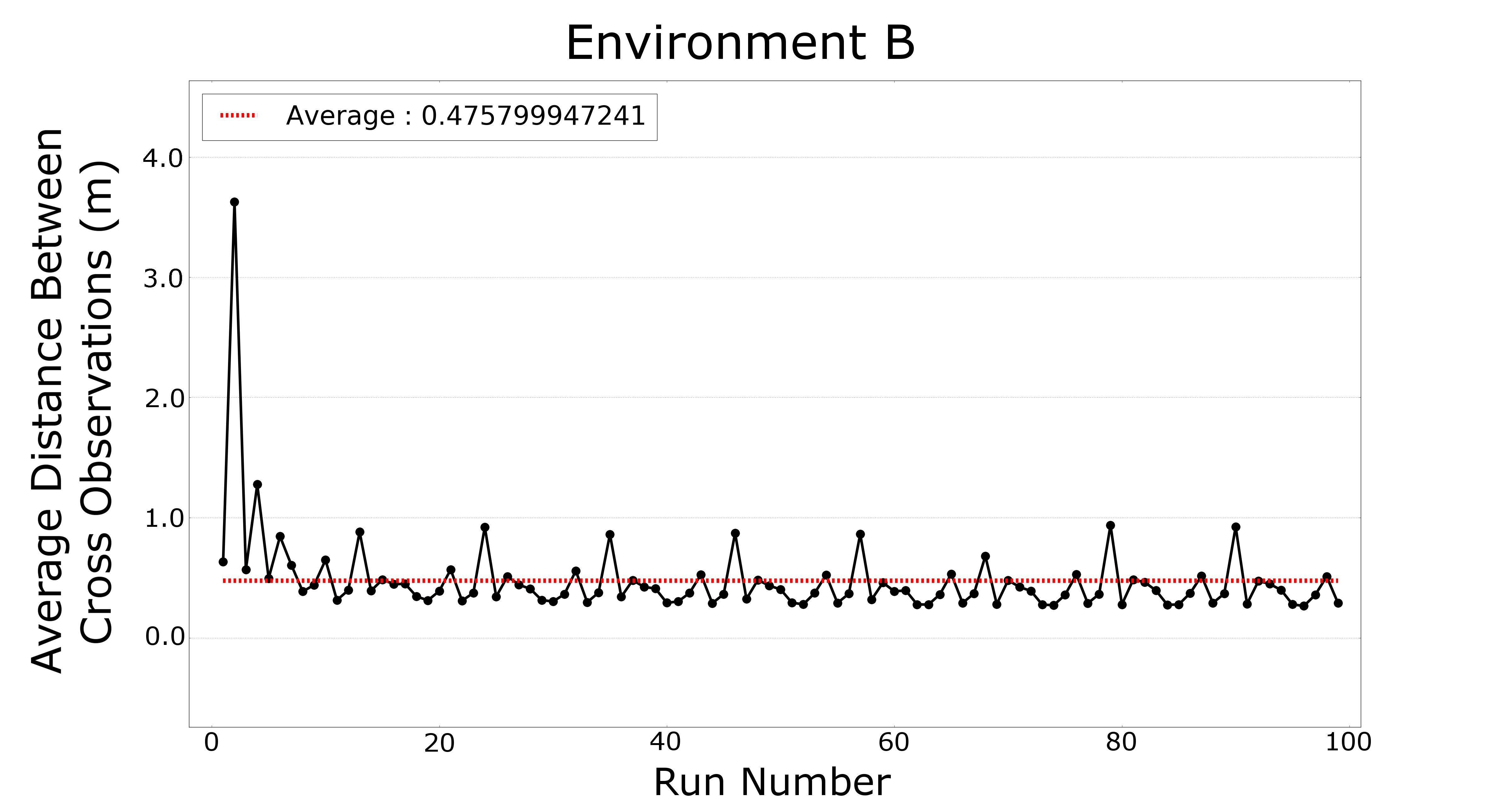}
            \label{fig:avg_dist_x_obs_nandan_house}
        \end{subfigure}
        ~
        \begin{subfigure}[b]{0.38\textwidth}
            \includegraphics[width=\textwidth]{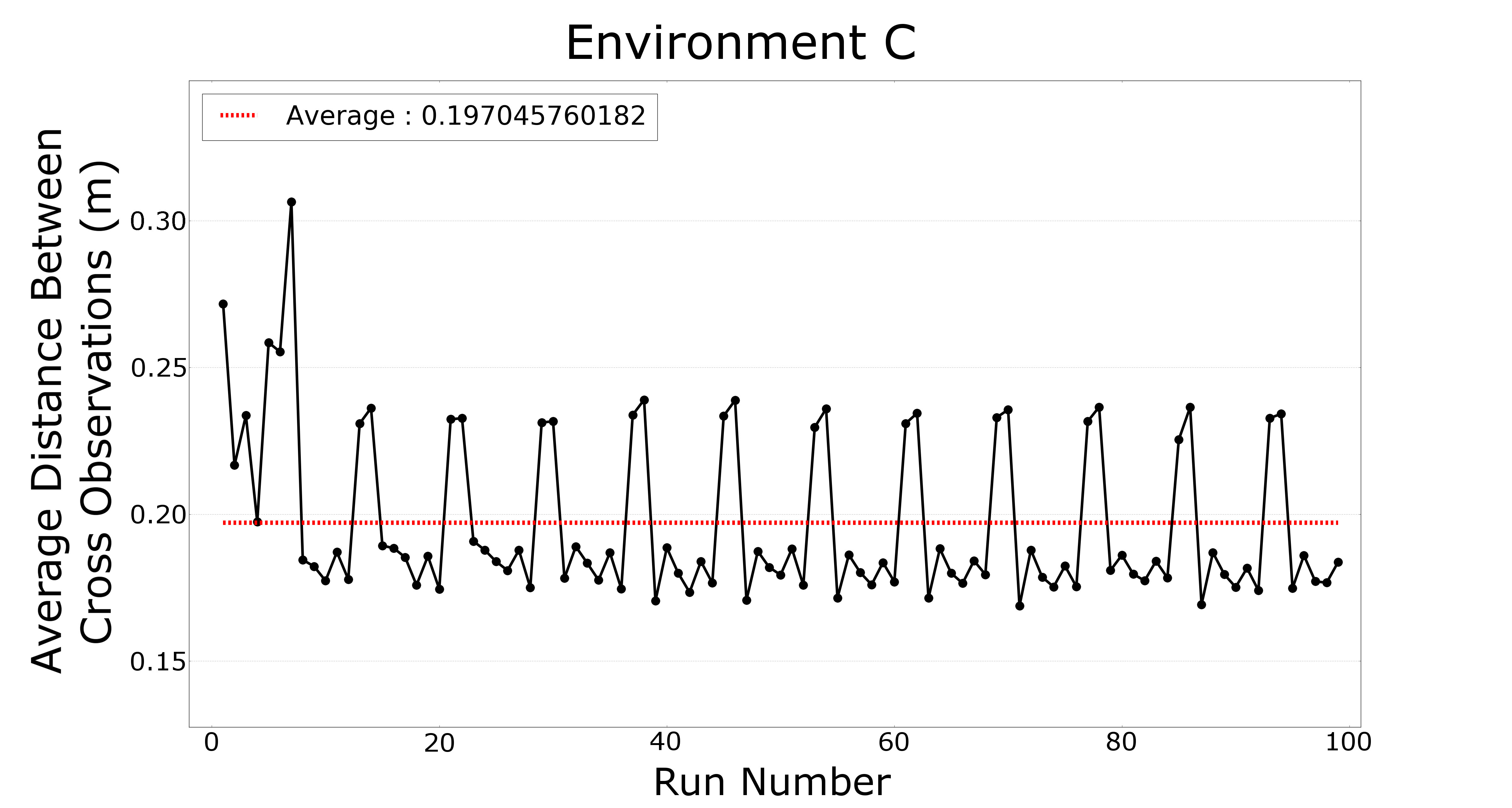}
            \label{fig:avg_dist_x_obs_pasadena_test_area_1}
        \end{subfigure}
        ~
        \begin{subfigure}[b]{0.38\textwidth}
            \includegraphics[width=\textwidth]{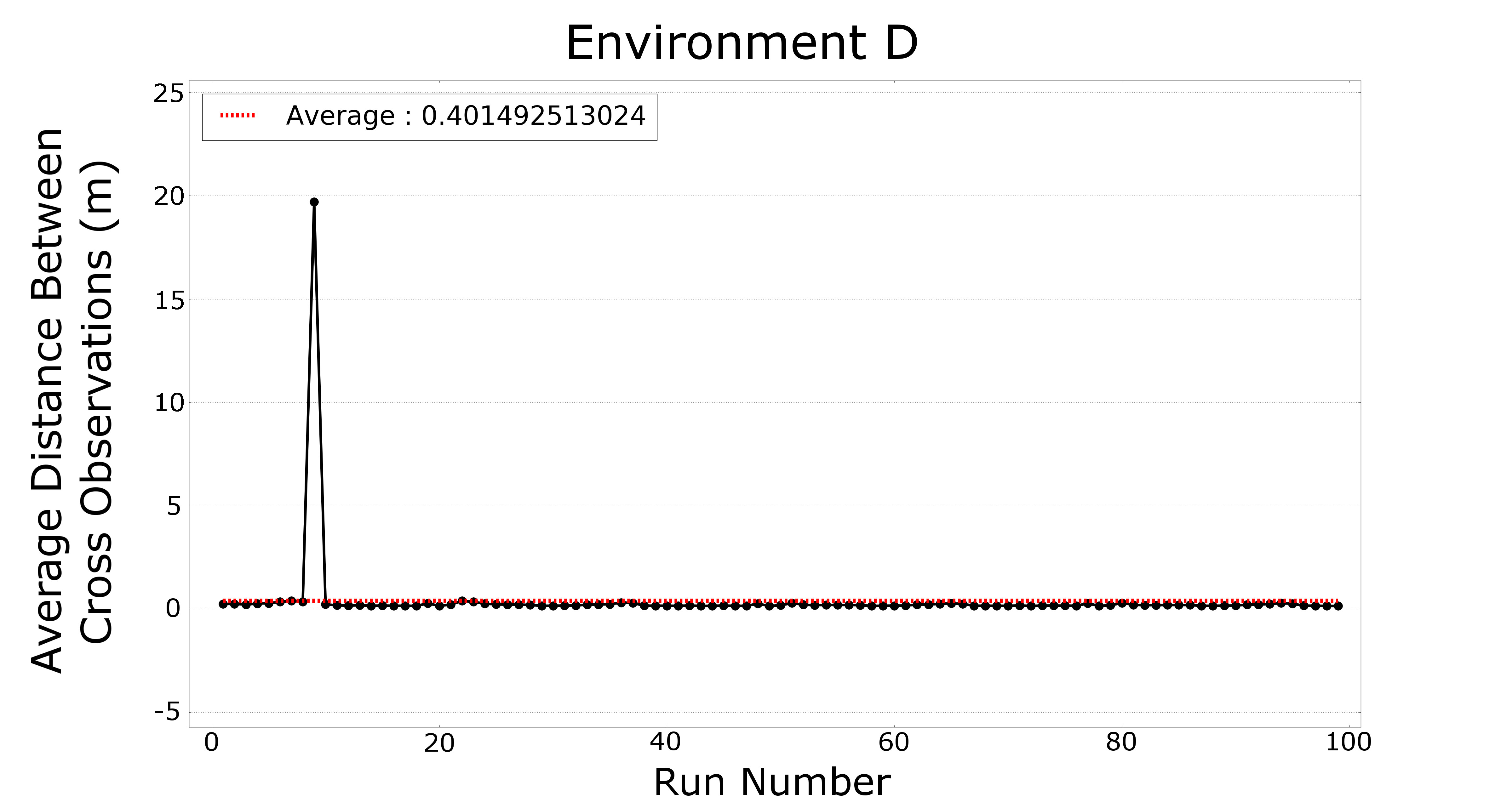}
            \label{fig:avg_dist_x_obs_pasadena_test_area_2}
        \end{subfigure}
    \caption{The average distance between cross observations, the lower the better. In the four environments, it starts off a little higher and then it makes observations from views from the loaded map every 0.5m. In environment D, the average distance between cross observations for the 9th run is very high because the lighting conditions were drastically different from the previous runs and there were only a handful of views created in earlier runs that the robot could observe in the 9th run. Once the robot created new views in that lighting condition, it didn't have any problems thereafter.}
    \label{fig:avg_dist_x_obs_all_envs}
\end{figure*}

% ($W_1 = 1.5, W_2 = 1, W_3 = 3$, score threshold of 0.25 times max\_score)
\begin{table}[t]
\centering
\vspace{5mm}
\captionof{table}{Nearest neighbor threshold selection} \label{tab:nn_thresh_selection_table}
\begin{tabular}{@{}cccc@{}}
\toprule
\textbf{NN\_THRESH} & \textbf{\begin{tabular}[c]{@{}c@{}}VOXEL\_PARAMS\\ (x, y, theta) {[}m, m, rad{]}\end{tabular}} & \textbf{\begin{tabular}[c]{@{}c@{}}Relocalization\\ Distance {[}m{]}\end{tabular}} & \textbf{\begin{tabular}[c]{@{}c@{}}Growth\\ Rate\end{tabular}} \\ \midrule
1                   & 1, 1, 1                                                                                            & 13.78                                                                              & 4.05                                                  \\
1                   & 1, 1, 2                                                                                            & 14.615                                                                             & 3.15                                                           \\
\textbf{1}                   & \textbf{2, 2, 1}                                                                                            & \textbf{9.349}                                                                              & \textbf{4.2}                                                            \\
1                   & 2, 2, 2                                                                                            & 11.256                                                                             & 3.95                                                           \\
\textbf{3}                   & \textbf{1, 1, 1}                                                                                            & \textbf{9.183}                                                                     & \textbf{4.45}                                                  \\
\textbf{3}                   & \textbf{1, 1, 2}                                                                                            & \textbf{11.583}                                                                    & \textbf{4.05}                                                  \\
3                   & 2, 2, 1                                                                                            & 9.417                                                                              & 3.55                                                           \\
3                   & 2, 2, 2                                                                                            & 10.409                                                                             & 4                                                              \\
5                   & 1, 1, 1                                                                                            & 11.527                                                                             & 5                                                              \\
\textbf{5}                   & \textbf{1, 1, 2}                                                                                            & \textbf{9.658}                                                                     & \textbf{4.4}                                                   \\
5                   & 2, 2, 1                                                                                            & 10.841                                                                             & 4                                                              \\
5                   & 2, 2, 2                                                                                            & 12.939                                                                             & 3.95                                                           \\ \bottomrule
\end{tabular}
\end{table}

We looked at the relocalization distance and the growth rate metrics to determine the appropriate nearest neighbor voxel size and threshold. In Table~\ref{tab:nn_thresh_selection_table}, the view score is obtained using $W_1 = 1.5$, $W_2 = 1$, $W_3 = 3$, and a view score threshold of 1.375. Nearest neighbor threshold is varied from 1 to 5, and the voxel size is varied from (1m, 1m, 1 rad) to (2m, 2m, 2 rad). The relocalization distance and the growth rates are compared with the values from using just the view score without the nearest neighbor constraint. The average relocalization distance is 11.78m and growth rate is 4.25 without using the nearest neighbor constraint. Growth rates within a tolerance of 0.2 (4.05 to 4.45) and a lower relocalization distance ($\le$ 11.78m) is used for selecting the parameters. Using these, we find (1, 2m, 2m, 1 rad), (3, 1m, 1m, 1 rad), (3, 1m, 1m, 2 rad), and (5, 1m, 1m, 2 rad) to be suitable nearest neighbor threshold and voxel size candidates.

\subsection{Lifelong mapping runs}
\label{sec:lifelong_mapping}

%After carefully selecting appropriate scoring weights, score threshold, nearest neighbor voxel parameters, and nearest neighbor threshold
We show the results of the proposed view management algorithm by running it on four different environments (see Fig.~\ref{fig:maps_with_views}) for 100 runs (see Fig.~\ref{fig:growth_all_envs}, \ref{fig:reloc_distance_all_envs},  \ref{fig:frac_x_observed_all_envs}, \ref{fig:avg_dist_x_obs_all_envs}). The weights used for computing the view score were (1.5, 1, 3) with a score threshold of 1.375, and a nearest neighbor voxel size of (1m, 1m, 2 rad) with a nearest neighbor threshold of 5. A minimum views threshold was set at 25, as the system can easily handle a low number of views and pruning is not necessary. The  periodicity that is seen in some of the plots is due to cycling of some of the logs during the sequential 100 runs. For these environments, and in general environments below 1000 $ft^2$, we have seen the total number of views usually stabilize within 300 with our algorithm. Without running our algorithm, the total number of views can go up to 1500 views increasing the CPU usage while observing views. The runtime of the algorithm on our mobile robot platform is around 300 ms with a system containing up to 500 views on a 1.2 GHz quad core cellphone grade processor. 

%The view management algorithm has been tested in real world user homes after being integrated into the Roomba i7 robot platform enabling robust lifelong mapping.

\section{Conclusions and Future Work}

In this paper we have addressed the problem of view management in a visual SLAM system, which arises during lifelong visual mapping on a resource-constrained mobile robot platform. We have presented an efficient and robust algorithm for deciding which views can be removed.  We have also devised formal criteria for measuring how well the robot is able to relocalize. We have then used them to demonstrate experimentally that our algorithm is capable of limiting the growth of the number of views in the SLAM system without compromising its ability to relocalize, despite changes in the appearance of the environment.

We have described the procedure we used for tuning the algorithm's parameters. One possible future research direction is to apply machine learning techniques to automate that process, and possibly improve performance further. %While this algorithm was designed with a monocular visual SLAM system in mind, it should be possible to apply it to systems using other sensors, such as RGB-D or stereo cameras.

\section{Acknowledgements}
%We would like to acknowledge the contributions of Manjunath Narayana, who has designed the performance metrics and benchmarks we used for testing and evaluating our approach. We would also like to thank Philip Fong, Renaud Moser, Martin Llofriu, and Emily Pittore for their advice, and their work on the overall system.

We would like to thank Philip Fong, Renaud Moser, Martin Llofriu, and Emily Pittore for their advice, and their work on the overall system.

\bibliographystyle{IEEEtran}
\bibliography{references}

\begin{thebibliography}{10}
\providecommand{\url}[1]{#1}
\csname url@samestyle\endcsname
\providecommand{\newblock}{\relax}
\providecommand{\bibinfo}[2]{#2}
\providecommand{\BIBentrySTDinterwordspacing}{\spaceskip=0pt\relax}
\providecommand{\BIBentryALTinterwordstretchfactor}{4}
\providecommand{\BIBentryALTinterwordspacing}{\spaceskip=\fontdimen2\font plus
\BIBentryALTinterwordstretchfactor\fontdimen3\font minus
  \fontdimen4\font\relax}
\providecommand{\BIBforeignlanguage}[2]{{%
\expandafter\ifx\csname l@#1\endcsname\relax
\typeout{** WARNING: IEEEtran.bst: No hyphenation pattern has been}%
\typeout{** loaded for the language `#1'. Using the pattern for}%
\typeout{** the default language instead.}%
\else
\language=\csname l@#1\endcsname
\fi
#2}}
\providecommand{\BIBdecl}{\relax}
\BIBdecl

\bibitem{Eade2010}
E.~Eade, P.~Fong, and M.~E. Munich, ``{Monocular graph SLAM with complexity
  reduction},'' \emph{IEEE IROS}, pp. 3017--3024, 2010.

\bibitem{konolige2010view}
K.~Konolige, J.~Bowman, J.~Chen, P.~Mihelich, M.~Calonder, V.~Lepetit, and
  P.~Fua, ``View-based maps,'' \emph{The International Journal of Robotics
  Research}, vol.~29, no.~8, pp. 941--957, 2010.

\bibitem{ava500}
``{iRobot}'s new {Ava} 500 puts robotics in heart of the enterprise,''
  \url{https://www.forbes.com/sites/jenniferhicks/2013/06/10/irobots-new-ava-500-puts-robotics-in-heart-of-the-enterprise/#3506cd972e1b},
  accessed: 2018-09-14.

\bibitem{konolige2009towards}
K.~Konolige and J.~Bowman, ``Towards lifelong visual maps,'' in \emph{IEEE
  IROS}, 2009, pp. 1156--1163.

\bibitem{poseGraph}
N.~S{\"u}nderhauf and P.~Protzel, ``Towards a robust back-end for pose graph
  slam,'' in \emph{IEEE ICRA}, May 2012, pp. 1254--1261.

\bibitem{factorGraphs}
\BIBentryALTinterwordspacing
F.~Dellaert and M.~Kaess, ``Factor graphs for robot perception,''
  \emph{Foundations and Trends® in Robotics}, vol.~6, no. 1-2, pp. 1--139,
  2017. [Online]. Available: \url{http://dx.doi.org/10.1561/2300000043}
\BIBentrySTDinterwordspacing

\bibitem{Duy2018}
D.-N. Ta, N.~Banerjee, S.~Eick, S.~Lenser, and M.~Munich, ``Fast nonlinear
  approximation of pose graph node marginalization,'' in \emph{IEEE ICRA}, May
  2018.

\bibitem{lowe-kd-trees}
M.~Muja and D.~G. Lowe, ``Fast approximate nearest neighbors with automatic
  algorithm configuration,'' in \emph{In VISAPP International Conference on
  Computer Vision Theory and Applications}, 2009, pp. 331--340.

\bibitem{hochdorfer}
S.~Hochdorfer, M.~Lutz, and C.~Schlegel, ``Lifelong localization of a mobile
  service-robot in everyday indoor environments using omnidirectional vision,''
  in \emph{{IEEE} International Conference on Technologies for Practical Robot
  Applications ({TEPRA})}, November 2009.

\bibitem{Hartmann}
W.~{Hartmann}, M.~{Havlena}, and K.~{Schindler}, ``Predicting matchability,''
  in \emph{IEEE CVPR}, June 2014, pp. 9--16.

\bibitem{buoncompagni2015saliency}
S.~Buoncompagni, D.~Maio, D.~Maltoni, and S.~Papi, ``Saliency-based keypoint
  selection for fast object detection and matching,'' \emph{Pattern Recognition
  Letters}, vol.~62, pp. 32--40, 2015.

\bibitem{erasingbadmemories}
M.~Dymczyk, T.~Schneider, I.~Gilitschenski, R.~Siegwart, and E.~Stumm,
  ``Erasing bad memories: agent-side summarization for long-term mapping,'' in
  \emph{IEEE IROS}, October 2016.

\bibitem{summary_maps_for_lifelong_visual_localization}
\BIBentryALTinterwordspacing
P.~M{\"u}hlfellner, M.~B{\"u}rki, M.~Bosse, W.~Derendarz, R.~Philippsen, and
  P.~Furgale, ``Summary maps for lifelong visual localization,'' \emph{Journal
  of Field Robotics}, vol.~33, no.~5, pp. 561--590, 2015. [Online]. Available:
  \url{https://onlinelibrary.wiley.com/doi/abs/10.1002/rob.21595}
\BIBentrySTDinterwordspacing

\bibitem{BuerkiAppearance}
M.~{B{\"u}rki}, I.~{Gilitschenski}, E.~{Stumm}, R.~{Siegwart}, and J.~{Nieto},
  ``Appearance-based landmark selection for efficient long-term visual
  localization,'' in \emph{IEEE IROS}, Oct 2016, pp. 4137--4143.

\bibitem{keepitbrief}
M.~{Dymczyk}, S.~{Lynen}, M.~{Bosse}, and R.~{Siegwart}, ``Keep it brief:
  Scalable creation of compressed localization maps,'' in \emph{IEEE IROS},
  Sep. 2015, pp. 2536--2542.

\bibitem{Churchill}
W.~Churchill and P.~Newman, ``Practice makes perfect? managing and leveraging
  visual experiences for lifelong navigation,'' in \emph{IEEE ICRA}, May 2012,
  pp. 4525--4532.

\bibitem{Churchilltwo}
\BIBentryALTinterwordspacing
------, ``Experience-based navigation for long-term localisation,'' \emph{The
  International Journal of Robotics Research}, vol.~32, no.~14, pp. 1645--1661,
  2013. [Online]. Available: \url{https://doi.org/10.1177/0278364913499193}
\BIBentrySTDinterwordspacing

\end{thebibliography}
\end{document}